\documentclass[sigconf,natbib,screen=true]{acmart}

\usepackage[skip=0pt]{caption}
\usepackage{multirow}
\usepackage{CJKutf8}
\usepackage[utf8]{inputenc}
\usepackage{array}
\usepackage{makecell}
\usepackage{graphicx}
\usepackage{amsmath}
\usepackage{url}
\usepackage{booktabs}
\usepackage{array}
\usepackage{algorithm}
\usepackage{algorithmic}
\usepackage{color}
\usepackage[inline]{enumitem}
\usepackage{todonotes}
\usepackage{ifthen}
\usepackage{xspace}
\usepackage{textcomp}
\usepackage{float}
\usepackage{colortbl}
\usepackage{subcaption}
\usepackage{multirow}
\usepackage{ctable} 
\usepackage{wrapfig}
\usepackage[normalem]{ulem}

\makeatletter
\newcommand{\thickhline}{%
    \noalign {\ifnum 0=`}\fi \hrule height 1pt
    \futurelet \reserved@a \@xhline
}
\makeatother

\copyrightyear{2021}
\acmYear{2021}
\setcopyright{acmlicensed}\acmConference[WSDM '21]{Proceedings of the Fourteenth ACM International Conference on Web Search and Data Mining}{March 8--12, 2021}{Virtual Event, Israel}
\acmBooktitle{Proceedings of the Fourteenth ACM International Conference on Web Search and Data Mining (WSDM '21), March 8--12, 2021, Virtual Event, Israel}
\acmPrice{15.00} 
\acmDOI{10.1145/3437963.3441804} 
\acmISBN{978-1-4503-8297-7/21/03}

\AtBeginDocument{%
\providecommand\BibTeX{{%
\normalfont B\kern-0.5em{\scshape i\kern-0.25em b}\kern-0.8em\TeX}}}

\author{Qintong Li}
\authornote{Work performed during an internship at Tencent AI Lab.}
\orcid{}
\affiliation{
 \institution{Shandong University}
}
\email{qintongli@mail.sdu.edu.cn}

\author{Piji Li}
\orcid{}
\affiliation{
 \institution{Tencent AI Lab}
}
\email{pijili@tencent.com}

\author{Xinyi Li}
\orcid{}
\affiliation{
 \institution{Peng Cheng Laboratory}  
}
\email{lixinyimichael@gmail.com}

\author{Zhaochun Ren}
\authornote{Zhaochun Ren is the corresponding author.}
\orcid{}
\affiliation{
 \institution{Shandong University}
}
\email{zhaochun.ren@sdu.edu.cn}

\author{Zhumin Chen}
\orcid{}
\affiliation{
 \institution{Shandong University}
}
\email{chenzhumin@sdu.edu.cn}

\author{Maarten de Rijke}
\orcid{0000-0002-1086-0202}
\affiliation{
 \institution{University of Amsterdam}   \& Ahold Delhaize
}
\email{m.derijke@uva.nl}

\title{Abstractive Opinion Tagging}
\allowdisplaybreaks

\begin{document}

\begin{abstract}
In e-commerce, \emph{opinion tags} refer to a ranked list of tags provided by the e-commerce platform that reflect characteristics of reviews of an item. 
To assist consumers to quickly grasp a large number of reviews about an item, opinion tags are increasingly being applied by e-commerce platforms.
Current mechanisms for generating opinion tags rely on either manual labelling or heuristic methods, which is time-consuming and ineffective. 
In this paper, we propose the \textit{abstractive opinion tagging} task, where systems have to automatically generate a ranked list of opinion tags that are based on, but need not occur in, a given set of user-generated reviews.  

The abstractive opinion tagging task comes with three main challenges: 
\begin{enumerate*}
\item the noisy nature of reviews;
\item the formal nature of opinion tags vs.\ the colloquial language usage in reviews; and 
\item the need to distinguish between different items with very similar aspects.
\end{enumerate*}
To address these challenges, we propose an abstractive opinion tagging framework, named AOT-Net, to generate a ranked list of opinion tags given a large number of reviews. 
First, a \textit{sentence-level salience estimation} component estimates each review's salience score. 
Next, a \textit{review clustering and ranking} component ranks reviews in two steps: first, reviews are grouped into clusters and ranked by cluster size; then, reviews within each cluster are ranked by their distance to the cluster center. 
Finally, given the ranked reviews, a \textit{rank-aware opinion tagging} component incorporates an alignment feature and alignment loss to generate a ranked list of opinion tags. 
To facilitate the study of this task, we create and release a large-scale dataset, called \textit{eComTag}, crawled from real-world e-commerce websites.
Extensive experiments conducted on the \textit{eComTag} dataset verify the effectiveness of the proposed AOT-Net in terms of various evaluation metrics.
\end{abstract}



\maketitle


\section{Introduction}

\begin{figure}[t]
  \begin{footnotesize}
  \begin{tabular}{@{~}p{8.2cm}@{~}} 
  \toprule
  \multicolumn{1}{c}{\textbf{Reviews of a hot-pot restaurant}}
  \\
  \midrule
  \begin{minipage}[t]{8.2cm}
  \vspace*{-.15cm}
  \begin{itemize}[leftmargin=*]
  \item $U_1$: The waitress was extremely attentive and even gave us a free fried man tou dessert that came with condensed milk for dipping...I love it!! 
  \item $U_2$: I was pleasantly surprised about how yummy the dish and the lamb were\ldots
  \item $U_3$: All in all; was a great experience and the service is really above and beyond.
  \item $U_4$: The restaurant guest is more, can be served quickly, our table was quickly dish bowl filled with. Overall cool experience. 
  \item $U_5$: The shrimp was fresh and the pork mixture was tasty.
  \item $U_6$: Fairly quick and polite service. It worth that price!
  \item  \ldots\ excellent, relaxed and cozy atmosphere,  and what can I say, satisfying.
  \item $U_N$: Food is delicious, reasonably priced\ldots Go here! you deserve it! 
  \vspace*{-.3cm}
  \end{itemize}
  \end{minipage}\\
  \multicolumn{1}{c}{} 
  \\
  \midrule
  \multicolumn{1}{c}{\textbf{Opinion tags}} 
  \\ 
  \midrule
  hospitable service (223), delicious food (165), value for money (104), comfortable environment (65), served quickly (14).
  \\
  \bottomrule
  \end{tabular}
  \end{footnotesize} 
  \caption{An example of a set of reviews and their corresponding opinion tags, where $U_N$ denotes the index of reviews, whereas the number behind each opinion tag reflects the number of reviews belonging to the tag.}
      \label{fig:exp}
\end{figure}

With the explosive growth of customer reviews in e-commerce scenarios,
many online platforms, such as Amazon\footnote{\url{https://www.amazon.com/}} and Alibaba\footnote{\url{https://www.alibaba.com/}}, provide opinion tags to enable potential buyers to make informed decisions without having to absorb large numbers of reviews.
As shown in Figure~\ref{fig:exp}, a sequence of opinion tags is a ranked list mined from a set of reviews that reflects different users' preferences towards certain aspects of items.
Many studies have focused on mining valuable information from reviews and shown promising results in various tasks, such as opinion summarization~\cite{Amplayo19, Lihr19, BrazinskasLT20, AmplayoL20, SuharaWAT20, li2017neural,li2019persona,CarmeliWSALLT20} and item description generation~\cite{novgorodov2019generating, EladGNKR19}. 
So far, however, no study seems to have developed opinion tagging methods to generate opinion tags that reflect diverse opinions of item aspects in a concise manner. 
In this paper, we propose the task of \textit{abstractive opinion tagging}, which aims to automatically generate a ranked list of opinion tags that stem from, but need not occur in, a given set of user-generated reviews.
This is an \emph{abstractive} rather than an \emph{extractive} opinion tagging task as the opinion tags do not need to occur in the review.

\begin{figure}[t]
  \includegraphics[width=\columnwidth]{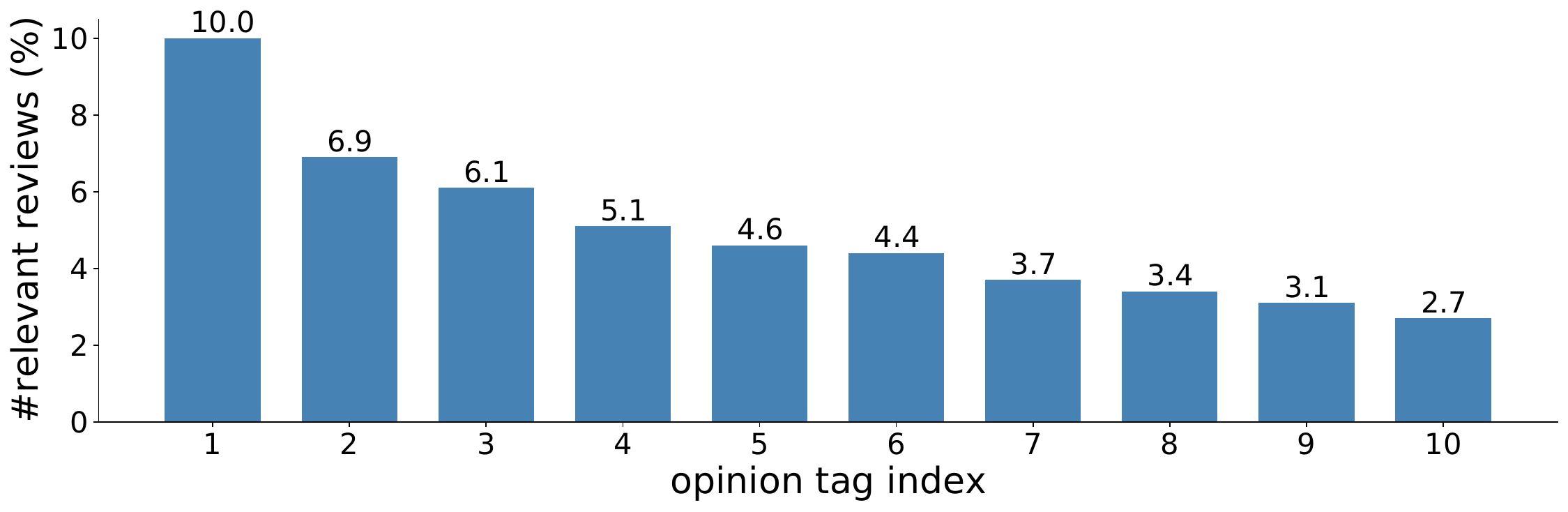}
  \caption{Fraction of relevant reviews per opinion tag in the \textit{eComTag} dataset.} 
  \Description{Reviews per opinion tag.}
  \label{fig:relevantreviews}
\end{figure}

To solve this new task, we face three challenges.
First, in reviews of e-commerce items, the noisy nature of the reviews inevitably makes it hard to identify salient item-related features~\citep{GaoRZZYY19,Lihr19}.
According to human annotations (on the eComTag dataset described below), we find that almost 59\% of review sentences are not item-related or do not contain opinions towards certain aspects.
Second, different reviewers have different ways of expressing themselves~\citep{tang2015user},
whereas the target opinion tags are usually in a more formal style.
The colloquial language usage of user reviews~\citep{WangLCKLS19} makes abstractive methods necessary to learn better review representations. 
Third, it is difficult to reflect the different perspectives on an item in an accurate and diverse manner.
Many e-commerce platforms \emph{rank} opinion tags to help customers distinguish between different items with very similar aspects.
In Figure~\ref{fig:relevantreviews}, we plot the number of relevant reviews on different ranked indexes of opinion tags in the \textit{eComTag} dataset (described below), which suggests a natural ranking for the tags when presented to users.

To address the challenges listed above, we design an abstractive framework, named AOT-Net, which consists of three components: 
\begin{enumerate*}
\item a \textit{sentence-level salience estimation} component that predicts a salience score for each review;
\item a \textit{review clustering and ranking} component that first groups reviews into clusters, which we refer to as ``opinion clusters,'' and ranks reviews by opinion cluster size; this component then ranks reviews within each cluster by their distance to the cluster center; and
\item a \textit{rank-aware opinion tagging} component that generates opinion tags with ranks.
\end{enumerate*}

\begin{figure}[!t]
  \centering
  \includegraphics[width=\linewidth]{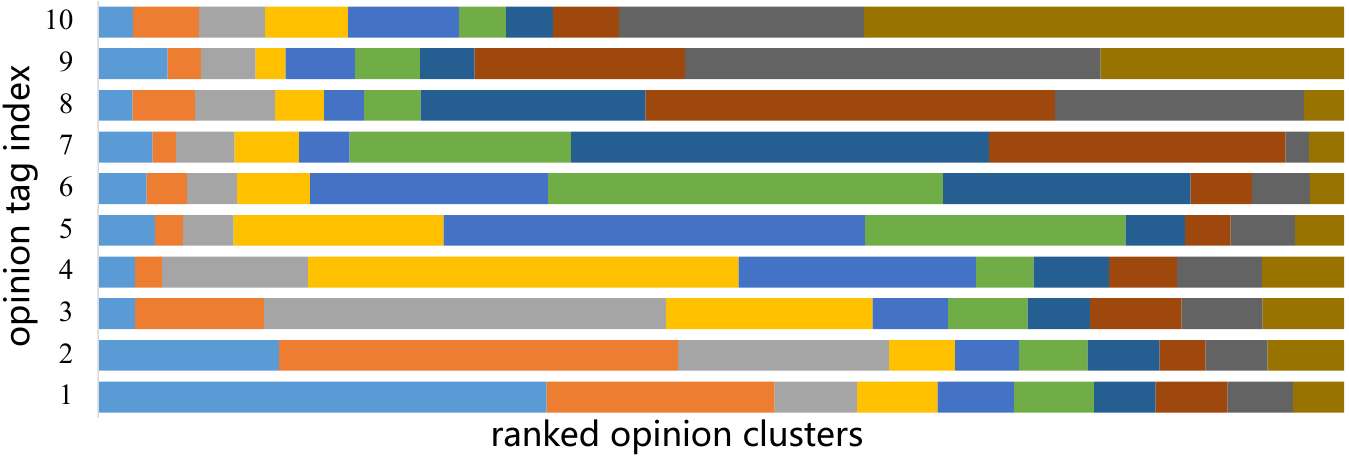}
\caption{The average semantic similarity between opinion tags and opinion clusters for all samples in the \textit{eComTag} dataset. Each color denotes an opinion cluster. The same opinion clusters share the same color across different opinion tags. The width of the color band denotes the degree of semantic similarity between an opinion tag and the corresponding opinion cluster. \textit{(Best viewed in color.)}}
  \Description{Semantic similarity.}
  \label{fig:rank_alignment}
\end{figure}

AOT-Net works in such a way that the ranks of the opinion clusters are correlated with the ranks of opinion tags. 
To see why this is potentially useful, we consider the eComTag dataset (crawled from e-commerce sites and consisting of items, reviews, and opinion tags), group reviews into opinion clusters and rank clusters by cluster size. 
We then compute the semantic similarity between the opinion clusters and the opinion tags, obtained by averaging the semantic similarity between an opinion tag and reviews in an opinion cluster. 
Figure~\ref{fig:rank_alignment} visualizes the average semantic similarity between $10$ opinion tags and $10$ opinion clusters for all item samples in the \textit{eComTag} dataset.
Clearly, the opinion tags have broader chunks with the opinion clusters at the same rank,  revealing an alignment between the ranked opinion tags and ranked opinion clusters. 
Furthermore, the opinion tags are semantically similar to the neighbors of the corresponding opinion clusters, which is reflected by the relatively wide color bands, e.g., the second opinion tag is semantically similar to the first opinion cluster and the third opinion cluster. 
The \textit{rank-aware opinion tagging} component of AOT-Net integrates two alignment strategies: 
\begin{enumerate*}
\item an opinion tag and its corresponding opinion clusters are placed at similar ranks; and
\item an \textit{alignment loss} explicitly encourages the model to focus on the aligned opinion clusters and ignore others.
\end{enumerate*}

To validate AOT-Net, we collect a new dataset, named \textit{eComTag}, from Chinese e-commerce websites, containing reviews and opinion tags for 50,068 items.
Our experiments based on the \emph{eComTag} dataset show that AOT-Net is capable of significantly improving the generation performance on abstractive opinion tagging over state-of-the-art baselines.

Our contributions can be summarized as follows:

\begin{itemize}[leftmargin=*]
\item We propose a new task, \textit{abstractive opinion tagging}, to generate opinion tags based on large volumes of item reviews. 
\item We propose an abstractive framework AOT-Net to generate opinion tags based on reviews for a given item. AOT-Net has a sentence-level salience estimation component and a review clustering and ranking component to highlight salient reviews and rank reviews by clustering. We propose a rank-aware opinion tagging component with two alignment strategies to generate ranked opinion tags.  
\item We collect a large-scale dataset, namely \textit{eComTag}, consisting of item reviews and opinion tags, to support research into abstractive opinion tagging. Experimental results conducted on the \textit{eComTag} dataset demonstrate the effectiveness of the AOT-Net framework.
\end{itemize}


\section{Related Work}
\label{section:related work}

Related work comes in two categories: keyphrase generation and opinion summarization.

\subsection{Keyphrase Generation}
A lot of research has been conducted on generating keyphrases to summarize various types of text such as tweets, news reports, research articles, etc.~\cite{MengZHHBC17, chen2019guided, liu2020keyphrase,ZhangLSZ18, WangLKLS19, WangLCKLS19, ray2019keyphrase, liu2020diverse}.
Early approaches to keyphrase generation extract important phrases from the document as the results.
Sequence tagging models have been applied to identify keyphrases~\cite{ZhangWGH16, luan-etal-2017-scientific, gollapalli2017incorporating}.
Retrieval-based approaches utilize a two-step pipeline to extract and rank candidate keyphrases~\cite{mihalcea2004textrank, medelyan2009human, wang2016ptr, LeNS16}.
\citet{SunTDDN19} adopt an extractive graph-based approach, which applies a point network to generate a set of diverse keyphrases.
Recently, abstractive approaches have also been explored.
\citet{meng2017deep} are the first to employ attention-based seq2seq framework with copy mechanism to conduct abstractive keyphrase generation.
\citet{ChanCWK19} propose a reinforcement learning approach for neural keyphrase generation that encourages a model to generate both sufficient and accurate keyphrases.
\citet{WangLCKLS19} propose a topic-aware neural keyphrase generation method to identify topic words. 

Unlike the work listed above, which only considers keyphrase generation for single document, we consider opinion tagging from multiple documents, that is, from all of the reviews for a given item.

\subsection{Opinion Summarization}

Opinion summarization has become an emerging research topic in recent years.
Early studies on opinion summarization focus on extracting salient sentences from the original review text~\citep{HuL04, CareniniNP06, LuZS09, GanesanZH10, XiongL14, AngelidisL18}: \citet{HuL04} identify item features mentioned in the reviews and then extract opinion sentences for the identified features. 
\citet{XiongL14} utilize unsupervised learning methods to extract review summaries by exploiting review helpfulness ratings.
\citet{AngelidisL18} present a weakly supervised neural framework for aspect-based opinion summarization by combining the tasks of aspect extracting and sentiment predicting.
Reflecting the most representative opinions from reviewers, many recent studies have shown that abstractive approaches are more appropriate for summarizing review text~\citep{CareniniCP13, GeraniMCNN14, FabbrizioSG14, LiWYZ19, Tay19}: 
\citet{GeraniMCNN14} utilize a template filling strategy to indirectly generate a review summary; 
\citet{WangL16} apply an attention-based encoder-decoder framework to generate an abstractive summary for opinionated documents.
The main objective of the above summarization approaches is to generate coherent sentences to summarize opinions.

In contrast, we propose the \textit{abstractive opinion tagging} task so as to generate opinion tags from a large number of user-generated reviews. 
In our scenario, opinion tags are more concise but without loss of essential information; they should help users comprehend reviews quickly and conveniently~\citep{li2017neural}.


\section{Problem Formulation} 
\label{sec:prel}

Before detailing our proposed method, AOT-Net, we first formulate the abstractive opinion tagging problem.
We use bold lowercase characters to denote vectors, and bold upper case characters to denote matrices. 
We write \textbf{W} and \textbf{b} for a projection matrix and a bias vector in a neural network layer, respectively.
Suppose that there are $M$ \emph{reviews} for a given item. 
We denote each review $X_i, 1 \leq i \leq M$ as a sequence of words, i.e., $X_i = [x_1, \ldots, x_{L_{x_i}}]$, where $L_{x_i}$ denotes the number of words in $X_i$.
In the same way, we assume that $N$ \emph{opinion tags} exist for a given item.
We denote each opinion tag $Y_j, 1 \leq j \leq N$ as a sequence of words, i.e., $Y_j = [y_1, \ldots, y_{L_{y_j}}]$, where $L_{y_j}$ refers to the number of words in $Y_j$.
Given a set of reviews $\mathcal{X}=\left \{X_1, X_2, \ldots, X_M\right \}$, the task of \emph{abstractive opinion tagging} is to generate a sequence of opinion tags $\mathcal{Y}=[Y_1, Y_2, \ldots, Y_N]$.


\section{Method}

\begin{figure*}[t]
  \centering
  \includegraphics[width=0.96\textwidth]{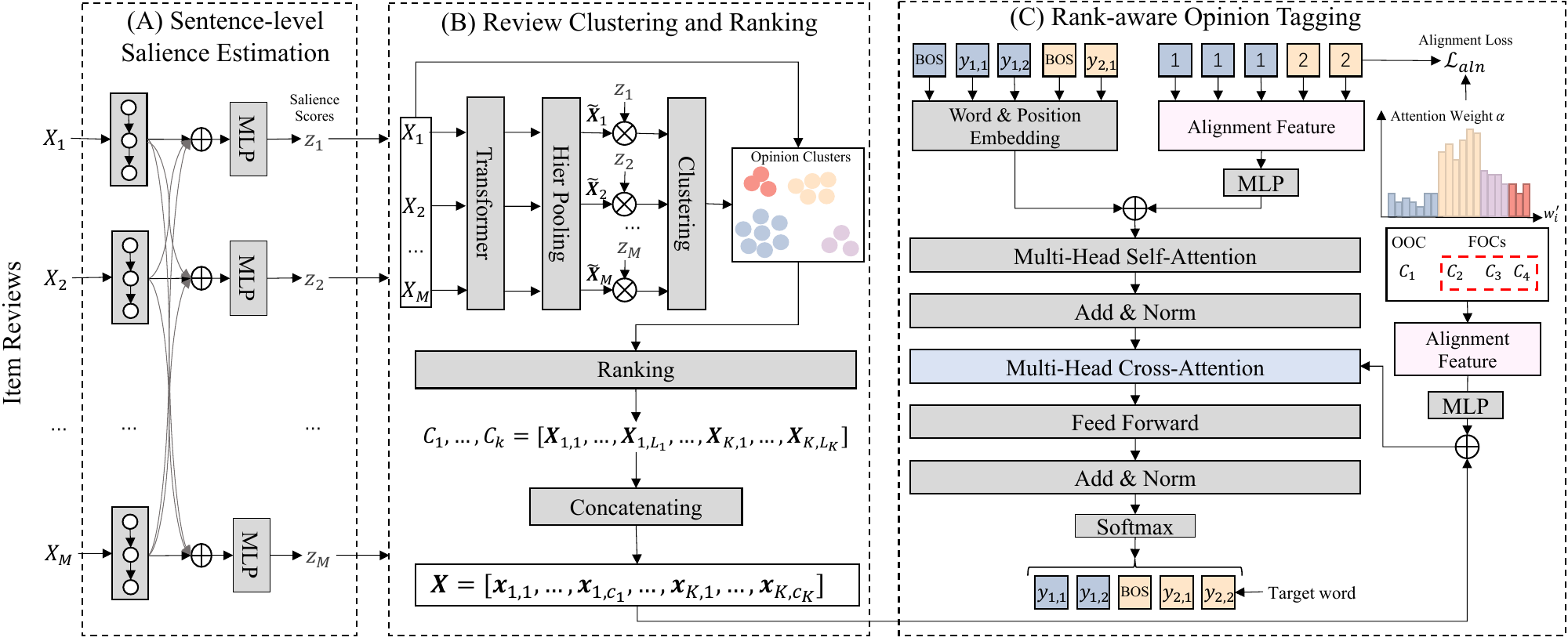}
  \caption{Our proposed framework AOT-Net for abstractive opinion tagging.}
  \Description{Model overview.}
  \label{fig:model}
\end{figure*}

\subsection{Overview} \label{sec:overview}

Before providing the details of AOT-Net, our proposed method for abstractive opinion tagging, we first provide an overview in Figure~\ref{fig:model}.
We divide AOT-Net into three main phases: 
\begin{enumerate*}[label=(\Alph*)]
\item sentence-level salience estimation; 
\item review clustering and ranking; and 
\item rank-aware opinion tagging.
\end{enumerate*}
For a set of reviews $\mathcal{X}$ about a given item, in phase A, we derive a salience score $z_i$ for each review $X_i \in \mathcal{X}$ to estimate its item-aware salience information.
In phase B, reviews are first encoded into vector representations and weighted by corresponding salience scores.
The weighted vector representations of reviews are clustered into $K$ opinion clusters $\{C_1, \ldots, C_K\}$ and ranked by cluster size.
Reviews within each cluster are ranked by their distance to the cluster center.
Then we flatten ranked reviews into word-level vector representations.  
In phase C, we use review representations to generate ranked opinion tags via two alignment constraints, i.e., alignment features and alignment loss.
We jointly learn all components in a multi-task learning framework.

\subsection{A: Sentence-level Salience Estimation} 
\label{sec:sen-est}

The aim of the sentence-level salience estimation component is to compute a salience score for each review $X_i \in \mathcal{X}$.
We design a \emph{sentence-level self-attention mechanism} to highlight item-related reviews and reduce noise.
First, the component reads each review sequence $X_i = [x_1, \ldots, x_{L_{x_i}}]$ and
uses a lookup table to convert each review word $x_p$ to a word embedding vector $\mathbf{x}_p \in \mathbb{R}^{d_e}$.
To incorporate the contextual information of the review text into the representation of each word, 
we feed each embedding vector $\mathbf{x}_p$ to a bi-directional Gated-Recurrent Unit (GRU)~\cite{ChoMGBBSB14} to learn a hidden representation $\mathbf{h}_p \in \mathbb{R}^{d}$.
More specifically, a bi-directional GRU consists of a forward GRU that reads the embedding sequence from $\mathbf{x}_1$ to $\mathbf{x}_{L_{x_i}}$
and a backward GRU that reads from $\mathbf{x}_{L_{x_i}}$ to $\mathbf{x}_1$:
\begin{align}
    \overrightarrow{\mathbf{h}}_p &= \text{GRU}_{f}(\mathbf{x}_p, \overrightarrow{\mathbf{h}}_{{p-1}}), \\
    \overleftarrow{\mathbf{h}}_p &= \text{GRU}_{b}(\mathbf{x}_p, \overleftarrow{\mathbf{h}}_{{p+1}}),
\end{align}
where $\overrightarrow{\mathbf{h}}_p \in \mathbb{R}^{d/2}$ and $\overleftarrow{\mathbf{h}}_p \in \mathbb{R}^{d/2}$ denote the hidden states of the forward $\text{GRU}_{f}$ and backward $\text{GRU}_{b}$, respectively.
We concatenate the last forward hidden state $\overrightarrow{\mathbf{h}}_{L_{x_i}}$ and last backward hidden state $\overleftarrow{\mathbf{h}}_1$ to form the hidden representation for review $X_i$,
i.e., $\mathbf{h}_{X_i}=[\overrightarrow{\mathbf{h}}_{L_{x_i}};\overleftarrow{\mathbf{h}}_1]$.

Next, we pass hidden representations of all reviews $\{\mathbf{h}_{X_i}\}_{i=1:M}$ to a self-attention layer to model more complex interactions among the reviews.
We propose a salience context vector $\textbf{c}_i$ for each review $X_i$ to denote the shared information from other reviews:
\begin{align}
  \mathbf{q}_{i} &= \mathbf{W}_q\mathbf{h}_{X_i}, \quad
    \mathbf{k}_{i} = \mathbf{W}_k\mathbf{h}_{X_i}, \quad 
    \mathbf{v}_{i} = \mathbf{W}_v\mathbf{h}_{X_i}, \\
  \mathbf{c}_{i} &= \sum_{i'=1}^{M}\frac{\text{exp}(\mathbf{q}_i^T\mathbf{k}_{i'})}{\sum_{o=1}^M\text{exp}(\mathbf{q}_i^T\mathbf{k}_{o})}\mathbf{v}_{i'} \text{,}  
\end{align}
where $\textbf{q}_i, \textbf{k}_i, \textbf{v}_i \in \mathbb{R}^{d\times d}$ refer to query, key, and value vectors, respectively. 
These vectors are linearly transformed from review hidden representation $\mathbf{h}_{X_i}$.
Then we apply a residual connection from the review hidden representation $\mathbf{h}_{X_i}$ to the salience context vector $\mathbf{c}_{i}$ and feed it to a two-layer feed-forward network with a ReLU as the activation function:
\begin{align}
  \mathbf{h}'_{X_i} &= \mathbf{W}_{s1}(\text{ReLU}(\mathbf{W}_{s2}(\mathbf{h}_{X_i} + \mathbf{c}_i))).
\end{align}
Given the context-enhanced review hidden representation $\mathbf{h}'_{X_i}$, we can derive the real-valued salience score $z_i$ for $X_i$:
\begin{align}
  z_i = \sigma(\mathbf{W}_s\mathbf{h}'_{X_i}+b_s),
\end{align}
where $\mathbf{W}_s \in \mathbb{R}^d$ and $b_s \in \mathbb{R}$. $\sigma(\cdot)$ is the sigmoid activation function.
The salience scores $\{s'_1, \ldots, s'_M\}$ serve as the salience weights of review representations for the later review clustering and ranking component.

In order to optimize the sentence-level salience estimating component, we manually label a binary \textit{salience label} $z^*_i \in \{0,1\}$ for each review $X_i$, where 1 denotes ``item-related'' whereas 0 denotes ``noisy''.
Then, sentence-level salience estimation component is trained by minimizing the cross-entropy loss function:
\begin{align}
  \mathcal{L}_{cla} &=  -\frac{1}{M}\sum_{i=1}^M z^*_i\log(z_i)+(1-z^*_i)\log(1-z_i).
\end{align}

\subsection{B: Review Clustering and Ranking} 
\label{sec:clu-est}

We propose a review clustering and ranking component to learn the ranks of reviews by grouping reviews into ranked opinion clusters,
which is the main prerequisite to accurately generate ranked opinion tags.

We use a standard transformer encoder~\cite{vaswani2017attention} to convert each review $X_i$ into vector representations.
Following \citet{vaswani2017attention}, we first map each word $x \in X_i$ into its vectorized representation $\textbf{x} \in \mathbb{R}^{d_e}$ using a word embedding layer and a positional embedding layer, as shown in the following equation:
\begin{align}
  \textbf{x} &= \text{Embed}(x) + \text{Pos}(x).
\end{align}
Then we use a transformer layer to encode global contextual information for words within $X_i$. 
\begin{align}
  \mathbf{g} &= \text{LayerNorm}(\mathbf{x}^{n-1} + \text{MHAtt}(\mathbf{x}^{n-1})), \\
  \mathbf{x}^{n} &=  \text{LayerNorm}(\mathbf{g} + \text{FFN}(\mathbf{g})),
\end{align}
where LayerNorm is the layer normalization proposed by~\citet{BaKH16}; MHAtt is the multi-head attention mechanism introduced by~\citet{vaswani2017attention}; FFN is a two-layer feed-forward network with ReLU as hidden activation function; and $n$ is the number of transformer block layers.
The word-level vector representations of review $X_i$ are $\mathbf{X}_i = [\mathbf{x}_1, \ldots, \mathbf{x}_{L_{x_i}}] = [\mathbf{x}^{n}_1, \ldots, \mathbf{x}^{n}_{L_{x_i}}]$.
To obtain the sentence representation $\tilde{\mathbf{X}}_i$ of review $X_i$, we perform a hierarchical pooling operation~\cite{HenaoLCSSWWMZ18} across its different words.
The hierarchical pooling mechanism can preserve word order information and has demonstrated superior performance over mean-pooling or max-pooling on many semantic analysis tasks~\cite{HenaoLCSSWWMZ18}.

To highlight the item-aware reviews and ignore noisy reviews, the sentence representations of reviews are first weighted by the corresponding salience scores, i.e., $\mathbf{\tilde{X}}'_i =  z_i \mathbf{\tilde{X}}_i$.
Then we apply the $k$-means~\citep{macqueen1967some} algorithm on $\{\mathbf{\tilde{X}}'_1, \ldots, \mathbf{\tilde{X}}'_M\}$ to group corresponding $\{\mathbf{X}_1, \ldots, \mathbf{X}_M\}$ into $K$ opinion clusters.\footnote{$K$ is manually assigned according to the number of reviews. If $M \leq 200$, $K= \lceil \frac{M}{20}\rceil$, otherwise, $K=20$.}
We rank opinion clusters from the largest (representing the highest number of reviews) to the smallest, denoted as $[C_1, \ldots, C_K]$.
For each cluster, we rank reviews from the nearest (representing the distance between review and cluster center) to the farthest.
Finally, we obtain a ranked list of reviews, represented as $[\mathbf{X}_{1,1}, \ldots, \mathbf{X}_{1,{L_{1}}}, \ldots, \mathbf{X}_{K,{1}}, \ldots, \mathbf{X}_{K,{L_{K}}}]$,
where $\mathbf{X}_{k,i}$ is the $i$-th review in the $k$-th opinion cluster and $L_k$ is the number of reviews in the $k$-th opinion cluster. 

We sequentially concatenate the vector representations of reviews in the ranked list and derive the final word-level representations, i.e., $\mathbf{X}=[\mathbf{x}_{1,{1}}, \ldots, \mathbf{x}_{1,{c_1}}, \ldots, \mathbf{x}_{K,{1}}, \ldots, \mathbf{x}_{K,{c_K}}]$.\footnote{In this paper, we only focus on the ranks of opinion clusters. We regard reviews or words in the same opinion cluster as equally important.}
Similarly, $x_{k,p}$ is the $p$-th word in the $k$-th opinion cluster and $c_k$ is the number of words in the $k$-th opinion cluster. 
Next, $\mathbf{X}$ will serve as the memory bank for the later rank-aware opinion tagging component.

\subsection{C: Rank-aware Opinion Tagging} 
\label{sec:dec}
Accurately generating opinion tags with ranks is challenging.
Therefore, we propose a rank-aware opinion tagging component to generate ranked opinion tags. 

In the training stage, we add a start token $\texttt{BOS}$ at the beginning of each opinion tag, i.e., $Y'_j=[\texttt{BOS};Y_j]$ where $[;]$ is the concatenation function.
Then we concatenate all opinion tags into a sequence of words: $Y = [Y'_1; \ldots; Y'_N] = [y_{1,0},\ldots, y_{1, L_{y_1}}, \ldots, y_{N,0}, \ldots, y_{N,L_{y_N}}]$, where $y_{j,q}$ is the $q$-th word of opinion tag $Y'_j$, $y_{j,0}$ is the $\texttt{BOS}$ token of the $j$-th opinion tag.
Our decoder, i.e., the rank-aware opinion tagging component, follows the transformer architecture~\cite{vaswani2017attention}.

From the data analysis in Figure~\ref{fig:rank_alignment}, we know that the ranks of opinion tags have a strong correlation with the ranks of opinion clusters.
The $j$-th opinion tag may pay attention to the $j$-th opinion cluster and its surrounding neighbors simultaneously. 
Therefore, for each opinion tag $Y'_j$, we hypothesize that the model needs to focus on the $F$ most related opinion clusters.\footnote{$F$ is a hyperparameter. $F$ opinion clusters mean the $j$-th opinion clusters and its surrounding neighbors.}
If an opinion cluster belongs to the $F$ focused opinion clusters, we call it a \emph{Focused Opinion Cluster} (FOC).
Otherwise, we call it an \emph{Outer Opinion Cluster} (OOC).
We design two alignment strategies between between \text{FOC}s and opinion tags, i.e., \textit{incorporating alignment features} and \textit{enforcing alignment loss}, which help to improve the generation of ranked opinion tags.

\paragraph{Alignment Feature}
Intuitively, the opinion tags and their \text{FOC}s are semantically similar in the vector space.
To help the model capture the alignment between opinion tags and their \text{FOC}s, we incorporate alignment features $\text{Aln}(\cdot)$ into the word-level representations of the opinion tags and opinion clusters. 
Formally, $\text{Aln}(\cdot)$ is a function that maps an integer into a vector.
Now, we explain how we will use it to represent the ranks of opinion tags and opinion clusters.
First, we incorporate the alignment feature into the representations of opinion tags.
The words in the $j$-th opinion tag have the same rank $j$ where $1 \leq j \leq N$. 
For each word $y_{j,q}$, the vectorized representation is the sum of the alignment feature $\text{Aln}(j)$, the word embedding $\text{Embed}(y_{j,q})$, and the positional embedding $\text{Pos}(y_{j,q})$: 
\begin{align}
  \mathbf{y}_{j,q} &=  \mathbf{W}_{rt} \text{Aln}(j) + \text{Embed}(y_{j,q}) + \text{Pos}(y_{j,q}),
\end{align}
where $\mathbf{W}_{rt} \in \mathbb{R}^{d_e \times d_e}$ is a trainable model parameter. 

\begin{figure}[t]
 \centering
 \includegraphics[width=0.35\textwidth]{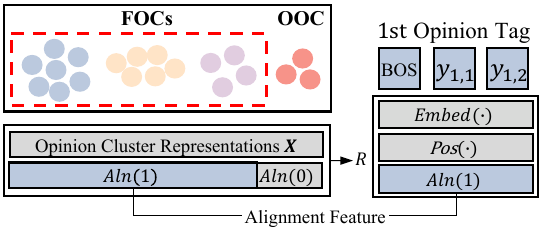}
 \caption{Representations attached with alignment features for opinion tags and opinion clusters.}
 \label{fig:dec_alignment}
 \vspace{-3mm}
\end{figure}

For the target $j$-th opinion tag, the ranks of \text{FOC}s are set to $j$ as well, while the ranks of \text{OOC}s are set to $0$.
Then we enhance the word vectors in $\textbf{X}$ with alignment features to capture the alignment between reviews and opinion tags:
\begin{align}
      \text{aln}_{x_{i,p}} = 
      \begin{cases}
        \text{Aln}(j), &  x_{i,p} \in \text{FOC}s\\
        \text{Aln}(0), & x_{i,p} \in \text{OOC}s.
      \end{cases}
\end{align}
Next, the alignment features $[\text{aln}_{x_{1,1}}, \ldots, \text{aln}_{x_{K,c_K}}]$ are added into the review representations $[\mathbf{x}_{1,1}, \ldots, \mathbf{x}_{K,c_K}]$ to obtain the align\-ment-enhanced representations  $\mathbf{R}=[r_{1,1},\ldots,r_{K,c_K}]$ for the words in opinion clusters:
\begin{align}
  \mathbf{r}_{i,p} &= \mathbf{W}_{rc} \text{aln}_{x_{i,p}} + \mathbf{x}_{i,p} \text{,}
\end{align}
where $\mathbf{W}_{rc} \in \mathbb{R}^{d_e}$ is a model parameter.
Figure~\ref{fig:dec_alignment}  summarizes the construction.

At each decoding step $q$ of the $j$-th opinion tag, the decoder reads the embeddings of the last prediction $\mathbf{y}_{j,q-1}$.
\begin{align}
  \mathbf{y}_{j,q} &= \text{transformer\_decoder}(\mathbf{y}_{j,q-1}),
\end{align}
where $\mathbf{y}_{j,q} \in \mathbb{R}^{d_e}$ is the target word representation.
Next, we introduce our decoder in detail.

To capture semantic and alignment information from the opinion clusters, a multi-head cross-attention $\text{MHAtt}$~\citep{vaswani2017attention} is applied to compute the attention score $[\alpha_{1,1}, \ldots, \alpha_{k,L_k}]$ between the last prediction $\mathbf{y}_{j,q-1}$ and $[\mathbf{r}_{1,1}, \ldots, \mathbf{r}_{k,c_k}]$:
\begin{align}
  \mathbf{u}_{j,q-1}^z &= \mathbf{W}_a^z \mathbf{y}_{j,q-1} \\
  \mathbf{k}_{i,p}^z &= \mathbf{W}_b^z \mathbf{r}_{i,p} \\
  \alpha_{i,p}^z &= \frac{\exp({\mathbf{u}_{j,q-1}^z}^T\mathbf{k}_{i,p}^z)}{\sum_{i'=1}^{K}\sum_{p'=1}^{c_{i'}}\exp({\mathbf{u}_{j,q-1}^z}^T\mathbf{k}_{i',p'}^z)},
\end{align}
where $\mathbf{u}_{j,q-1}^z \in \mathbb{R}^{d_h}, \mathbf{k}_{i,p}^z \in \mathbb{R}^{d_h}$ are query and key vectors that are linearly transformed from $\mathbf{y}_{j,q-1}$ and $\mathbf{r}_{i,p}$ as in~\citep{vaswani2017attention};
$z \in \{1, \ldots, n_h\}$ indicates the $z$-th head among $n_h$ heads; $d_h = d_e/n_h$ is the dimension of each head. 

The attention scores $[\alpha_{i,j}^1,\ldots,\alpha_{i,j}^{n_h}]$ are then used to compute an aggregated vector $\mathbf{c}_{j,q}$ for target word $y_{j,q}$: 
\begin{align}
  c_{j,q} = [\sum_{i=1}^{K}\sum_{j=1}^{c_i}\alpha_{i,j}^1\mathbf{r}_{i,j}; \ldots; \sum_{i=1}^{K}\sum_{j=1}^{c_i}\alpha_{i,j}^{n_h}\mathbf{r}_{i,j}].
\end{align}
Then we feed the last word representation $\mathbf{y}_{j,q-1}$ and vector $\mathbf{c}_{j,q}$ to a two-layer feed-forward network with a ReLU as the activation function and a highway layer normalization on top:
\begin{align}
  \mathbf{s}_{j,q-1} &= \text{LayerNorm}(\mathbf{y}^{n-1}_{j,q-1} + \mathbf{c}_{j,q}) \\
  \mathbf{y}^{n}_{j,q-1} &=  \text{LayerNorm}(\mathbf{s}_{j,q-1} + \text{FFN}(\mathbf{s}_{j,q-1})),
\end{align}
where $\mathbf{y}_{j,q} = \mathbf{y}^{n}_{j,q-1}$ is the target word representation.
After that, we use $\mathbf{y}_{j,q}$ to compute a probability distribution over the words in a predefined vocabulary $\mathcal{V}$,
as shown in the following equation:
\begin{align}
  P_\mathcal{V}(y_{j,q} \mid [y_{1,0},\ldots,y_{j,q-1}],\mathcal{X}) &= \text{softmax}(\mathbf{W}_v\mathbf{y}_{j,q} + \mathbf{b}_v),
\end{align}
where $\mathbf{W_v} \in \mathbb{R}^{|\mathcal{V}|\times d_e}$, $\mathbf{b}_v \in \mathbb{R}^{|\mathcal{V}|}$ are trainable parameters.
To enable our model to generate out-of-vocabulary (OOV) words, we adopt the copy mechanism~\citep{SeeLM17}
to predict OOV words by directly copying words from the opinion clusters.
We first compute a soft gate $p_{gen}\in[0,1]$ between generating a word from the predefined vocabulary $\mathcal{V}$ and copying a word from the input reviews $\mathcal{X}$: 
\begin{align} 
  p_{gen} &= \sigma(\mathbf{W_g}\mathbf{y}_{p,q} + b_g),
\end{align}
where $\mathbf{W_g} \in \mathbb{R}^{d_e}$ and $b_g \in \mathbb{R}$ are trainable parameters. 
Finally, we can derive the final next-word probability distribution $P(y_{j,q})$:
\begin{align}
  \alpha_{i,p} &= \frac{\sum_z^{n_h}\alpha_{i,p}^z}{n_h},\\
  P(y_{j,q}) &= p_{gen}P_\mathcal{V}(y_{j,q}) + (1-p_{gen})\sum_{i,p:x_{i,p}=y_{j,q}}\alpha_{i,p},
\end{align}
where we use $P(y_{j,q})$ to denote $P(y_{j,q}\mid [y_{1,0},\ldots,y_{j,q-1}],\mathcal{X})$ for brevity.
We use the negative log-likelihood of the ground-truth words $y^{*}_{j,q}$ as the generation loss function:
\begin{align}
  \mathcal{L}_{gen} &= -\sum_{j=1}^{N}\sum_{q=1}^{L_{y_j}} \log P(y^{*}_{j,q} \mid [\mathbf{y}^{*}_{1,0},\ldots,\mathbf{y}^{*}_{j,q-1}],\mathcal{X}).
\end{align}

\paragraph{Alignment Loss}
During the generation of the $j$-th opinion tag, we enforce an alignment loss to help locate \text{FOC}s accurately.
The model is explicitly taught to focus on \text{FOC}s and ignore \text{OOC}s in the attention $\alpha_{i,p}$ via the following loss:
\begin{align}
 \mbox{}\hspace*{-1.5mm}
  \mathcal{L}_{aln} = -\log\left(\frac{\sum_{i,p:x_{i,p} \in \text{FOC}s}\alpha_{i,p}}{\sum_{i,p}\alpha_{i,p}}\right) + \log\left(\frac{\sum_{i,p:x_{i,p} \in \text{OOC}s} \alpha_{i,p}}{\sum_{i,p}\alpha_{i,p}}\right)\!.
\end{align}

\subsection{Multi-task Training Objective}
We adopt a multi-task learning framework to jointly minimize the salience classification loss, alignment loss, and generation loss. 
The objective function is:
\begin{align}
  \mathcal{L} = \lambda_1 \mathcal{L}_{cla} + \lambda_2 \mathcal{L}_{aln} + \lambda_3 \mathcal{L}_{gen},
\end{align}
where $\lambda_1,\lambda_2,\lambda_3$ are hyper-parameters that control the weights of these three losses. 
We set $\lambda_1=\lambda_2=\lambda_3=1$.
Thus, each component of our joint model can be trained end-to-end.
 

\section{Experimental Setup}

We set up experiments to compare AOT-Net against a number of relevant baselines. 
We are interested in the overall performance of AOT-Net and in understanding the effectiveness of the salience estimation and ranking alignment.

\subsection{Experiments}
We report on five experiments.
First, we compare AOT-Net against a number of baselines to assess its overall performance. 
Then we conduct ablation studies to analyze the influence of different components in AOT-Net as follows:
\begin{enumerate*}[label=(\roman*)]
  \item \textbf{w/o SSE} is AOT-Net without the sentence-level salience estimating component (SSE). 
  \item \textbf{w/o RCR} is AOT-Net without the review clustering and ranking component (RCR).
  \item \textbf{w/o AF} is AOT-Net without alignment feature (AF).
  \item \textbf{w/o AL} is AOT-Net without alignment loss (AL).
\end{enumerate*}
To further explore the effectiveness of the sentence-level self-attention mechanism in SSE, we consider \textbf{AOT-RNN}, the method only considers BiGRU in salience score prediction; whereas we write \textbf{AOT-Embed} for the method that employs MLP to replace BiGRU in SSE. 
Fourth, we analyze the performance of AOT-Net for different sizes of \text{FOC}s.
Lastly, we provide a case study about abstractive opinion tagging.

\subsection{Baselines}
We compare AOT-Net with the following methods:
\begin{enumerate*}[label=(\roman*)]
\item \textbf{TF-IDF} is an extractive approach that selects the important words as summary based on term frequency and inverse document frequency;
\item \textbf{TextRank}~\cite{mihalcea2004textrank} is an unsupervised algorithm based on weighted-graphs;
\item \textbf{RNN} is a sequence to sequence model with attention implemented by bi-directional GRU layer~\cite{ChoMGBBSB14};
\item \textbf{PG-Net}~\cite{SeeLM17} is a classical opinion summarization model based on the encoder-decoder framework with attention and copy mechanisms; and 
\item the \textbf{Transformer}~\cite{vaswani2017attention} is a Transformer-based encoder-decoder model with a copy mechanism, which is a strong baseline widely-adopt in opinion summarization.
\end{enumerate*}

\subsection{The \emph{eComTag} Dataset}
Since there is no available opinion tagging dataset, we build a new one, named \textit{eComTag} from several Chinese e-commerce websites.
We collect nearly 112k items, sampling from different domains, including Cosmetic~($37.43\%$), Electronics~($29.51\%$), Books~($10.57\%$), Entertainment~($8.23\%$), Food~($7.62\%$), Sports~($3.96\%$), Clothes~($3.16\%$), Medical~($1.62\%$), and Furniture~($0.33\%$). For each domain, there are a set of reviews and a list of opinion tags.
Since reviewers may comment on multiple aspects, e.g., ``The dim sum tasted extremely fresh, and the price was quite reasonable!'', we split each review into sentences by punctuation. 
We use a sentence to denote a review in our paper.
Then, we remove samples where the number of opinion tags is smaller than 4 or the number of reviews is fewer than 50. 
Finally, we construct \textit{eComTag} with 50,068 item samples.
Users may write reviews arbitrarily, which results in many meaningless expressions, such as ``Love, love, LOVE this space!'', and ``come with my boyfriend.''
To teach our model to distinguish these noisy sentences, we annotate each review with a binary salience label via human judgment.
If a review is item-related, we label the review as 1, otherwise 0.
The salience labels are the supervision signals for the sentence-level salience estimating component.

Finally, each item sample consists of a set of reviews, a set of corresponding salience labels, and a sequence of opinion tags.
For text preprocessing, we tokenize texts using the Jieba toolkit\footnote{\url{https://github.com/fxsjy/jieba}} and maintain a 50k vocabulary. 
In \textit{eComTag}, about 30\% of samples have more than $1024$ words in reviews.
We randomly split the dataset into training/validation/test sets with 8:1:1 ratio. 
The statistics are shown in Table~\ref{tab:data_stats}. 
Specially, we define the \textit{present tag}~(Pr) as the exact tag that appears in reviews and \textit{absent tag}~(Ab) as the tag unseen in reviews.
The proportion of absent tags is close to 75\%, which further proves the necessity to apply abstractive methods on opinion tagging.

\subsection{Evaluation Metrics}
We employ two information retrieval metrics to evaluate the opinion tag generation: the macro F$@{k}$ score and the normalized discounted cumulative gain (NDCG$@{k}$) score.
Both are widely used to measure word overlap~\cite{SunTDDN19,ChanCK20}.
To measure diversity of the generated opinion tags,
we adopt the Distinct-${2}$ score~\citep{LiGBGD16} and a macro Unique-$N$ score.
We compute Unique-$N$ as follows, $\text{Unique-}N=\sum_{i=1}^{T}N_i/T$, 
where $N_i$ is the number of distinct opinion tags in the $i$-th sample.
Moreover, we design two metrics, Exact Rank Match (ERM) and Fuzzy Rank Match (FRM), to evaluate the rank accuracy of opinion tags. 
ERM is defined as the one-to-one exact match proportion between true tags and predicted tags.
Inspired by the Embedding Score~\cite{LiuLSNCP16}, FRM first maps predicted tags and the corresponding true tags into the same vector space, and then computes the average cosine similarity between their vector representations.

\begin{table}[!t]
\caption{Statistics of the \textit{eComTag} dataset. ``\textit{Pr}'' and ``\textit{Ab}'' denote the proportion of present tags and absent tags respectively. ``\textit{MTN}'' and ``\textit{MTL}'' indicate max tag number and max tag length  for a sample respectively.}
  \begin{tabular}{l c c c c c c}
    \toprule 
    Data & \em{Sample} & \em{Pr} & \em{Ab} & \em{MTN} & \em{MTL} \\  
    \midrule
    Training & 40,162 & 24.1\% & 75.9\% & 19 & 40\\  
    Validation & \phantom{0}4,953 & 24.3\% & 75.7\% & 20 & 39\\ 
    Test & \phantom{0}4,953 & 24.1\% & 75.9\% & 14 & 32\\ 
    \bottomrule
  \end{tabular}
\label{tab:data_stats}
 \vspace*{-.5\baselineskip}
\end{table}

\subsection{Implementation Details}
We adopt the Adam~\cite{KingmaB14} optimizer with settings \{$\beta_1 = 0.9$,$\beta_2 = 0.999$,$\epsilon = 10^{-8}$,$lr = 10^{-4}$\} and we vary the learning rate following~\citet{vaswani2017attention}.
We add dropout~\cite{SrivastavaHKSS14} with keeping rate 0.8 and label smoothing~\cite{SzegedyVISW16} with smoothing factor 0.1. 
We use the Tencent AI Lab Chinese Embeddings\footnote{\url{https://ai.tencent.com/ailab/nlp/embedding.html}} for initialization of the word embedding layers. 
The rest of the parameters are randomly initialized.
The dimensions of the alignment feature, word embedding layers and positional embedding layers are set to 200.
We set the batch size to 16 and use the validation loss for early stopping. 
When inference, we set the maximum decoding step as 50.
We use a bidirectional GRU~\cite{ChoMGBBSB14} with 2 layers to implement the sentence-level salience estimating component.
All RNN-based models have 256 hidden units.
All transformer-based models have 300 hidden units; the feed-forward hidden size is set to 50 for all layers.
We set the $F$ in Section~\ref{sec:dec} to $3$~($3$ is the number of \emph{Focused Opinion Cluster}s at each decoding step) to ensure the target tag token have enough relevant reviews to reference and avoid introducing too much interference information simultaneously.
AOT-Net was trained on a single Tesla V100 GPU and is implemented using PyTorch.
All hyperparameters and models are selected on the validation set and the results are reported on the test set.

\begin{table*}[t]
\renewcommand\arraystretch{1.0}
  \centering
  \caption{Evaluation results on the \textit{eComTag} dataset. Results in bold are leading results in terms of the corresponding metric.} 
  \begin{tabular}{l cc cc cc cc c }
    \toprule
    & & & & & & & \multicolumn{2}{c}{Distinct-2} 
    \\
    \cmidrule{8-9}
    {\textbf{Models}} &
    {F$_{1}@5$} &
    {F$_{1}@10$} &
    {NDCG@5} &
    {NDCG@10} &
    {ERM} &
    {FRM} &
    Micro & 
    Macro & 
    Unique-N
    \\
    \midrule
    TF-IDF & 0.0039 & 0.0038 & 0.0168 & 0.0169 & 0.16 & 0.19 & -- & -- & -- \\
    TextRank & 0.0019 & 0.0018 & 0.0091 & 0.0097 & 0.06 & 0.21 & -- & -- & -- \\
    RNN & 0.2895 & 0.2753 & 0.7383 & 0.7701 & 0.17 & 0.44 & 0.60 & 62.19 & 7.447 \\
    PG-Net & 0.3138 & 0.2896 & \textbf{0.7600} & 0.8009 & 0.19 & 0.44 & 1.33 & 65.78 & 6.798 \\
    Transformer & 0.2833 & 0.2756 & 0.6916 & 0.7483 & 0.25 & 0.59 & 1.57 & 89.23 & 8.851 \\
    \midrule
    \textbf{AOT-Net} & \textbf{0.3529} & \textbf{0.3492} & 0.7473 & \textbf{0.8045} & \textbf{0.31} & \textbf{0.64} & 1.30 & \textbf{94.35} & 8.953 \\
    w/o SSE & 0.2930 & 0.2822 & 0.7022 & 0.7563 & 0.25 & 0.61 & 1.11 & 91.85 & 8.957 \\
    w/o RCR & 0.3434 & 0.3370 & 0.7353 & 0.7913 & 0.31 & 0.63 & \textbf{1.77} & 92.94 & 8.935 \\
    w/o AF & 0.3141 & 0.3056 & 0.7194 & 0.7768 & 0.28 & 0.62 & 1.21 & 93.23 & \textbf{8.997} \\
    w/o AL & 0.3406 & 0.3336 & 0.7322 & 0.7857 & 0.30 & 0.64 & 1.30 & 93.11 & 8.926 \\
    \toprule
  \end{tabular}

  \label{tab:auto_result}
\end{table*}


\section{Results and Analysis}
Overall evaluation results on generating opinion tags are listed in Table~\ref{tab:auto_result}.
We find that all the abstractive models significantly outperform all the traditional extractive baselines. Thus we conclude that informal and colloquial nature of user-generated reviews make item-related features indiscernible using the unsupervised extraction methods. 
As expected, we also find that PG-Net significantly outperforms RNN, which implies that the copying mechanism is useful for opinion summarization.
We can see AOT-Net significantly outperforms baseline Transformer in terms of all metrics and achieves the best performance for most metrics.
The results of our ablation studies are shown in the lower part of the Table~\ref{tab:auto_result}. 
We observe that after removing the SSE component, performances of AOT-Net in terms of most metrics drops obviously.
If we do not rank and group reviews into opinion clusters before decoding (i.e., w/o RCR), although the diversity metric has a slight increase, the rank accuracy of AOT-Net decreases as we anticipated. 
We also find that after removing alignment feature or alignment loss mechanisms in the decoder, the performance of both retrieval and rank accuracy metrics (i.e., $F_1$ and ERM) degrades.
We will conduct a detailed analysis of the individual components in the following sections.

\subsection{Salience Estimation Analysis} 
\begin{figure}[h]
  \centering
  \includegraphics[clip,trim=2mm 0mm 1mm 0mm,width=\linewidth]{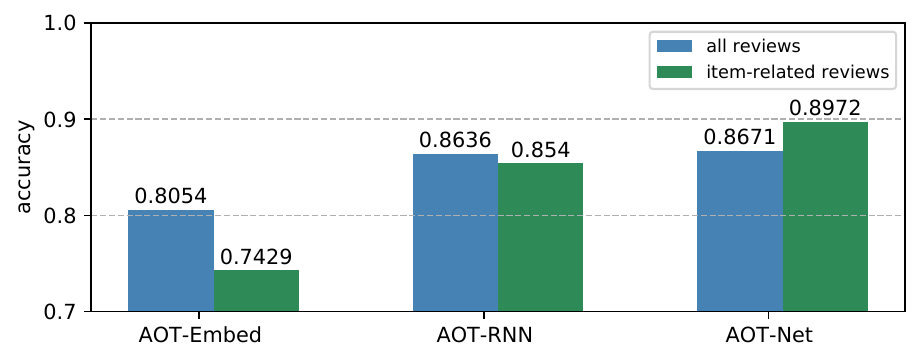}
  \caption{Accuracy values for sentence-level salience estimation.}
  \label{fig:cla_type}
  \vspace*{-.5\baselineskip}
\end{figure}

As shown in Table~\ref{tab:auto_result}, AOT-Net achieves a $17.1$\% and $2.72$\% increase over ``w/o SSE'' in terms of Micro-Distinct-2 and Macro-Distinct-2, respectively. Similar improvements can be observed for other metrics.
This demonstrates that the predicted salience scores help AOT-Net to focus on more valuable reviews. 
To verify the effectiveness of SSE with more details, in Figure~\ref{fig:cla_type} we list the accuracy scores of AOT-Embed, AOT-RNN, and AOT-Net for sentence-level salience estimation.
We find that AOT-Net outperforms both AOT-Embed and AOT-RNN, which verifies the effectiveness of BiGRU and self-attention mechanisms.
AOT-RNN achieves $7.2$\% increase over AOT-Embed in terms of accuracy for all reviews, which verifies the advantage of BiGRU in representing review. 
In terms of accuracy, we find that AOT-Net gives a $0.4$\% and $5.1$\% increase over AOT-RNN for all reviews and item-related reviews, respectively.
This indicates that AOT-Net benefits from self-attention mechanisms, which capture the shared information to distinguish item-related reviews.

\subsection{Number of FOCs}
To evaluate the effect of the number of FOCs on the performance of rank-aware opinion tagging, 
we examine the performance of AOT-Net with different values of $F$ (see Section~\ref{sec:dec}) in terms of ERM, FRM, and Distinct-2, respectively.
As shown in Table~\ref{tab:window_size}, $F=1$ significantly decreases the model rank and diversity performance.
This suggests that only focusing on a single opinion cluster ignores many related reviews.
We also find that when $F=5$, the performance of AOT-Net slightly decreases; AOT-Net achieves the best performance in terms of all metrics when $F=3$.
Hence, we infer that $F$ is a trade-off between focusing on relevant reviews and removing irrelevant noise.

\begin{table}[t]
  \caption{Performance on different numbers ($F$) of \text{FOC}s.}  
    \begin{tabular}{l c c c}
      \toprule 
      \bf Models & ERM & FRM & Macro Distinct-2 \\
      \midrule
      $F$=1 & 0.30 & 0.62 & 93.24 \\
      $F$=3 & 0.31 & 0.64 & 94.35 \\
      $F$=5 & 0.30 & 0.63 & 93.32 \\
      \bottomrule
    \end{tabular}
  \label{tab:window_size}
  \vspace*{-.5\baselineskip}
  \end{table}

\begin{figure}[h]
  \centering
  \includegraphics[clip,trim=2mm 0mm 1mm 0mm,width=0.98\linewidth]{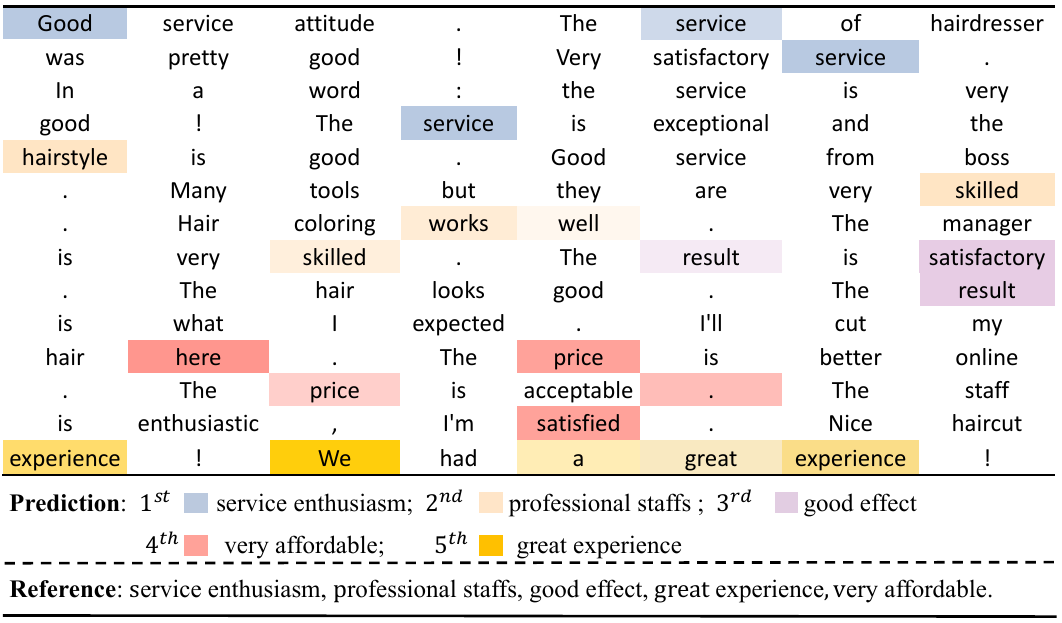}
  \caption{The transition of reviews attention distribution between tags computed by AOT-Net. Different colors correspond to different tags. We only depict $top@5$ attention probabilities for brevity. (Best viewed in color).}
  \Description{An example of cross-attention transit.}
  \label{fig:transit}
  \vspace*{-.5\baselineskip}
\end{figure}

\subsection{Case Study}

{Figure~\ref{fig:transit} shows an example illustrating the 5 highest attention weights $\alpha_{i,p}$ during the generation of opinion tags.}
We see that the model transits its focus smoothly from the first review sentence to later review sentences.   
Sometimes, the model may focus on the same review for two opinion tags, such as ``The \underline{service}~($1$st tag) is good and \underline{hairstyle}~($2$nd tag) is good.'' 
The first three predicted tags were completely accurate.
However, the transformer only predicted the first tag accurately. 
To validate the effectiveness of the alignment loss, 
we calculate $\sum_{i,p:x_{i,p}\in \text{FOCs}}$ and $\sum_{i,p:x_{i,p} \in \text{OOCs}}$ for all items in the test set.
Results show that $\sum_{i,p:x_{i,p}\in \text{FOCs}}$ and $\sum_{i,p:x_{i,p} \in \text{OOCs}}$ in AOT-Net are $0.7304$ and $0.2696$ on average.
However, for AOT-Net w/o AL, $\sum_{i,p:x_{i,p}\in \text{FOCs}}$ and $\sum_{i,p:x_{i,p}\in \text{OOCs}}$ are $0.6509$ and $0.3491$, respectively.
This phenomenon indicates that the alignment feature by itself is not enough to force AOT-Net to focus on FOCs.
In particular, we compare the ranks of opinion tags generated by the transformer and AOT-Net respectively. 
For the transformer, items where the first 3 opinion tags have accurate ranks only account for near 6.11\%, while for AOT-Net, the proportion can reach 9.68\%.
Thus, we conclude that the alignment feature and alignment loss in AOT-Net are helpful to capture the ranks of the opinion tags.



\section{Conclusion and Future Work}
In this paper, we have proposed the \textit{abstractive opinion tagging} task, which aims to automatically generate a ranked list of opinion tags from a large number of reviews.
We have proposed a rank-aware abstractive opinion tagging framework~(AOT-Net) that includes a sentence-level salience estimating component, a review clustering and ranking component, and a rank-aware opinion tagging component.
To validate the effectiveness of AOT-Net, we conduct extensive experiments on a newly collected real-world dataset, \textit{eComTag}. 
Experiments show that AOT-Net achieves state-of-the-art performance on the abstractive opinion tagging task.
AOT-Net has two main advantages over previous work. On the one hand, it generates more concise opinion tags; on the other hand, the ranked lists of generated opinion tags help users distinguish products with very similar aspects. 
Our work provides a plausible solution to greatly reduce human annotation costs for online e-commerce opinion tagging.
Although we focused mostly on e-commerce portals, our methods are also broadly applicable to other settings with opinionated content, such as microblogs. 

Limitations of our work include its low efficiency and coarse-grained salience estimation.
As to our future work, we will adopt the deep clustering network instead of $k$-means algorithm.
Also, pre-trained language models could provide more power to enhance our sentence salience estimation.
Few-shot learning for handling unbalanced review distributions among different domains could be another direction.
It will be also interesting to explore user interactions with AOT-Net to generate personalized opinion tags in the future.

\section*{Code and Data}
The source code and dataset used in this paper are available at \url{https://github.com/qtli/AOT}.

\begin{acks}
We thank our reviewers for valuable feedback.
This work was supported by the National Key R\&D Program of China with grant No. 2020YFB1406704, the Natural Science Foundation of China (61972234, 61902219, 61672324, 61672322, 62072279), the Key Scientific and Technological Innovation Program of Shandong Province (2019JZZY010129), the Tencent AI Lab Rhino-Bird Focused Research Program (JR201932), the Fundamental Research Funds of Shandong University, the Foundation of State Key Laboratory of Cognitive Intelligence, iFLYTEK, P.R. China (COGOSC-20190003).
All content represents the opinion of the authors, which is not necessarily shared or endorsed by their respective employers and/or sponsors.
\end{acks}

\bibliographystyle{ACM-Reference-Format}
\small
\bibliography{references}


\begin{thebibliography}{53}


\ifx \showCODEN    \undefined \def \showCODEN     #1{\unskip}     \fi
\ifx \showDOI      \undefined \def \showDOI       #1{#1}\fi
\ifx \showISBNx    \undefined \def \showISBNx     #1{\unskip}     \fi
\ifx \showISBNxiii \undefined \def \showISBNxiii  #1{\unskip}     \fi
\ifx \showISSN     \undefined \def \showISSN      #1{\unskip}     \fi
\ifx \showLCCN     \undefined \def \showLCCN      #1{\unskip}     \fi
\ifx \shownote     \undefined \def \shownote      #1{#1}          \fi
\ifx \showarticletitle \undefined \def \showarticletitle #1{#1}   \fi
\ifx \showURL      \undefined \def \showURL       {\relax}        \fi
\providecommand\bibfield[2]{#2}
\providecommand\bibinfo[2]{#2}
\providecommand\natexlab[1]{#1}
\providecommand\showeprint[2][]{arXiv:#2}

\bibitem[\protect\citeauthoryear{Amplayo and Lapata}{Amplayo and
  Lapata}{2019}]%
        {Amplayo19}
\bibfield{author}{\bibinfo{person}{Reinald~Kim Amplayo} {and}
  \bibinfo{person}{Mirella Lapata}.} \bibinfo{year}{2019}\natexlab{}.
\newblock \showarticletitle{Informative and Controllable Opinion
  Summarization}.
\newblock \bibinfo{journal}{\emph{CoRR}}  \bibinfo{volume}{abs/1909.02322}
  (\bibinfo{year}{2019}).
\newblock


\bibitem[\protect\citeauthoryear{Amplayo and Lapata}{Amplayo and
  Lapata}{2020}]%
        {AmplayoL20}
\bibfield{author}{\bibinfo{person}{Reinald~Kim Amplayo} {and}
  \bibinfo{person}{Mirella Lapata}.} \bibinfo{year}{2020}\natexlab{}.
\newblock \showarticletitle{Unsupervised Opinion Summarization with Noising and
  Denoising}. In \bibinfo{booktitle}{\emph{ACL}}. \bibinfo{pages}{1934--1945}.
\newblock


\bibitem[\protect\citeauthoryear{Angelidis and Lapata}{Angelidis and
  Lapata}{2018}]%
        {AngelidisL18}
\bibfield{author}{\bibinfo{person}{Stefanos Angelidis} {and}
  \bibinfo{person}{Mirella Lapata}.} \bibinfo{year}{2018}\natexlab{}.
\newblock \showarticletitle{Summarizing Opinions: Aspect Extraction Meets
  Sentiment Prediction and They Are Both Weakly Supervised}. In
  \bibinfo{booktitle}{\emph{EMNLP}}. \bibinfo{pages}{3675--3686}.
\newblock


\bibitem[\protect\citeauthoryear{Ba, Kiros, and Hinton}{Ba
  et~al\mbox{.}}{2016}]%
        {BaKH16}
\bibfield{author}{\bibinfo{person}{Lei~Jimmy Ba}, \bibinfo{person}{Jamie~Ryan
  Kiros}, {and} \bibinfo{person}{Geoffrey~E. Hinton}.}
  \bibinfo{year}{2016}\natexlab{}.
\newblock \showarticletitle{Layer Normalization}.
\newblock \bibinfo{journal}{\emph{CoRR}}  \bibinfo{volume}{abs/1607.06450}
  (\bibinfo{year}{2016}).
\newblock


\bibitem[\protect\citeauthoryear{Brazinskas, Lapata, and Titov}{Brazinskas
  et~al\mbox{.}}{2020}]%
        {BrazinskasLT20}
\bibfield{author}{\bibinfo{person}{Arthur Brazinskas}, \bibinfo{person}{Mirella
  Lapata}, {and} \bibinfo{person}{Ivan Titov}.}
  \bibinfo{year}{2020}\natexlab{}.
\newblock \showarticletitle{Unsupervised Opinion Summarization as
  Copycat-Review Generation}. In \bibinfo{booktitle}{\emph{ACL}}.
  \bibinfo{pages}{5151--5169}.
\newblock


\bibitem[\protect\citeauthoryear{Carenini, Cheung, and Pauls}{Carenini
  et~al\mbox{.}}{2013}]%
        {CareniniCP13}
\bibfield{author}{\bibinfo{person}{Giuseppe Carenini}, \bibinfo{person}{Jackie
  Chi~Kit Cheung}, {and} \bibinfo{person}{Adam Pauls}.}
  \bibinfo{year}{2013}\natexlab{}.
\newblock \showarticletitle{Multi-Document Summarization of Evaluative Text}.
\newblock \bibinfo{journal}{\emph{Comput. Intell.}} \bibinfo{volume}{29},
  \bibinfo{number}{4} (\bibinfo{year}{2013}), \bibinfo{pages}{545--576}.
\newblock


\bibitem[\protect\citeauthoryear{Carenini, Ng, and Pauls}{Carenini
  et~al\mbox{.}}{2006}]%
        {CareniniNP06}
\bibfield{author}{\bibinfo{person}{Giuseppe Carenini},
  \bibinfo{person}{Raymond~T. Ng}, {and} \bibinfo{person}{Adam Pauls}.}
  \bibinfo{year}{2006}\natexlab{}.
\newblock \showarticletitle{Multi-Document Summarization of Evaluative Text}.
  In \bibinfo{booktitle}{\emph{EACL}}. \bibinfo{publisher}{The Association for
  Computer Linguistics}.
\newblock


\bibitem[\protect\citeauthoryear{Carmeli, Wang, Suhara, Angelidis, Li, Li, and
  Tan}{Carmeli et~al\mbox{.}}{2020}]%
        {CarmeliWSALLT20}
\bibfield{author}{\bibinfo{person}{Nofar Carmeli}, \bibinfo{person}{Xiaolan
  Wang}, \bibinfo{person}{Yoshihiko Suhara}, \bibinfo{person}{Stefanos
  Angelidis}, \bibinfo{person}{Yuliang Li}, \bibinfo{person}{Jinfeng Li}, {and}
  \bibinfo{person}{Wang{-}Chiew Tan}.} \bibinfo{year}{2020}\natexlab{}.
\newblock \showarticletitle{ExplainIt: Explainable Review Summarization with
  Opinion Causality Graphs}.
\newblock \bibinfo{journal}{\emph{CoRR}}  \bibinfo{volume}{abs/2006.00119}
  (\bibinfo{year}{2020}).
\newblock


\bibitem[\protect\citeauthoryear{Chan, Chen, and King}{Chan
  et~al\mbox{.}}{2020}]%
        {ChanCK20}
\bibfield{author}{\bibinfo{person}{Hou~Pong Chan}, \bibinfo{person}{Wang Chen},
  {and} \bibinfo{person}{Irwin King}.} \bibinfo{year}{2020}\natexlab{}.
\newblock \showarticletitle{A Unified Dual-view Model for Review Summarization
  and Sentiment Classification with Inconsistency Loss}. In
  \bibinfo{booktitle}{\emph{SIGIR}}. \bibinfo{pages}{1191--1200}.
\newblock


\bibitem[\protect\citeauthoryear{Chan, Chen, Wang, and King}{Chan
  et~al\mbox{.}}{2019}]%
        {ChanCWK19}
\bibfield{author}{\bibinfo{person}{Hou~Pong Chan}, \bibinfo{person}{Wang Chen},
  \bibinfo{person}{Lu Wang}, {and} \bibinfo{person}{Irwin King}.}
  \bibinfo{year}{2019}\natexlab{}.
\newblock \showarticletitle{Neural Keyphrase Generation via Reinforcement
  Learning with Adaptive Rewards}. In \bibinfo{booktitle}{\emph{ACL}}.
  \bibinfo{pages}{2163--2174}.
\newblock


\bibitem[\protect\citeauthoryear{Chen, Gao, Zhang, King, and Lyu}{Chen
  et~al\mbox{.}}{2019}]%
        {chen2019guided}
\bibfield{author}{\bibinfo{person}{Wang Chen}, \bibinfo{person}{Yifan Gao},
  \bibinfo{person}{Jiani Zhang}, \bibinfo{person}{Irwin King}, {and}
  \bibinfo{person}{Michael~R Lyu}.} \bibinfo{year}{2019}\natexlab{}.
\newblock \showarticletitle{Title-Guided Encoding for Keyphrase Generation}. In
  \bibinfo{booktitle}{\emph{AAAI}}, Vol.~\bibinfo{volume}{33}.
  \bibinfo{pages}{6268--6275}.
\newblock


\bibitem[\protect\citeauthoryear{Cho, van Merrienboer, G{\"{u}}l{\c{c}}ehre,
  Bahdanau, Bougares, Schwenk, and Bengio}{Cho et~al\mbox{.}}{2014}]%
        {ChoMGBBSB14}
\bibfield{author}{\bibinfo{person}{Kyunghyun Cho}, \bibinfo{person}{Bart van
  Merrienboer}, \bibinfo{person}{{\c{C}}aglar G{\"{u}}l{\c{c}}ehre},
  \bibinfo{person}{Dzmitry Bahdanau}, \bibinfo{person}{Fethi Bougares},
  \bibinfo{person}{Holger Schwenk}, {and} \bibinfo{person}{Yoshua Bengio}.}
  \bibinfo{year}{2014}\natexlab{}.
\newblock \showarticletitle{Learning Phrase Representations using {RNN}
  Encoder-Decoder for Statistical Machine Translation}. In
  \bibinfo{booktitle}{\emph{EMNLP}}. \bibinfo{pages}{1724--1734}.
\newblock


\bibitem[\protect\citeauthoryear{Elad, Guy, Novgorodov, Kimelfeld, and
  Radinsky}{Elad et~al\mbox{.}}{2019}]%
        {EladGNKR19}
\bibfield{author}{\bibinfo{person}{Guy Elad}, \bibinfo{person}{Ido Guy},
  \bibinfo{person}{Slava Novgorodov}, \bibinfo{person}{Benny Kimelfeld}, {and}
  \bibinfo{person}{Kira Radinsky}.} \bibinfo{year}{2019}\natexlab{}.
\newblock \showarticletitle{Learning to Generate Personalized Product
  Descriptions}. In \bibinfo{booktitle}{\emph{CIKM}}.
  \bibinfo{pages}{389--398}.
\newblock


\bibitem[\protect\citeauthoryear{Fabbrizio, Stent, and Gaizauskas}{Fabbrizio
  et~al\mbox{.}}{2014}]%
        {FabbrizioSG14}
\bibfield{author}{\bibinfo{person}{Giuseppe~Di Fabbrizio},
  \bibinfo{person}{Amanda Stent}, {and} \bibinfo{person}{Robert~J.
  Gaizauskas}.} \bibinfo{year}{2014}\natexlab{}.
\newblock \showarticletitle{A Hybrid Approach to Multi-document Summarization
  of Opinions in Reviews}. In \bibinfo{booktitle}{\emph{INLG}}.
  \bibinfo{pages}{54--63}.
\newblock


\bibitem[\protect\citeauthoryear{Ganesan, Zhai, and Han}{Ganesan
  et~al\mbox{.}}{2010}]%
        {GanesanZH10}
\bibfield{author}{\bibinfo{person}{Kavita Ganesan}, \bibinfo{person}{ChengXiang
  Zhai}, {and} \bibinfo{person}{Jiawei Han}.} \bibinfo{year}{2010}\natexlab{}.
\newblock \showarticletitle{Opinosis: {A} Graph Based Approach to Abstractive
  Summarization of Highly Redundant Opinions}. In
  \bibinfo{booktitle}{\emph{COLING}}. \bibinfo{pages}{340--348}.
\newblock


\bibitem[\protect\citeauthoryear{Gao, Ren, Zhao, Zhao, Yin, and Yan}{Gao
  et~al\mbox{.}}{2019}]%
        {GaoRZZYY19}
\bibfield{author}{\bibinfo{person}{Shen Gao}, \bibinfo{person}{Zhaochun Ren},
  \bibinfo{person}{Yihong~Eric Zhao}, \bibinfo{person}{Dongyan Zhao},
  \bibinfo{person}{Dawei Yin}, {and} \bibinfo{person}{Rui Yan}.}
  \bibinfo{year}{2019}\natexlab{}.
\newblock \showarticletitle{Product-Aware Answer Generation in E-Commerce
  Question-Answering}. In \bibinfo{booktitle}{\emph{WSDM}}.
  \bibinfo{pages}{429--437}.
\newblock


\bibitem[\protect\citeauthoryear{Gerani, Mehdad, Carenini, Ng, and
  Nejat}{Gerani et~al\mbox{.}}{2014}]%
        {GeraniMCNN14}
\bibfield{author}{\bibinfo{person}{Shima Gerani}, \bibinfo{person}{Yashar
  Mehdad}, \bibinfo{person}{Giuseppe Carenini}, \bibinfo{person}{Raymond~T.
  Ng}, {and} \bibinfo{person}{Bita Nejat}.} \bibinfo{year}{2014}\natexlab{}.
\newblock \showarticletitle{Abstractive Summarization of Product Reviews Using
  Discourse Structure}. In \bibinfo{booktitle}{\emph{EMNLP}}.
  \bibinfo{pages}{1602--1613}.
\newblock


\bibitem[\protect\citeauthoryear{Gollapalli, Li, and Yang}{Gollapalli
  et~al\mbox{.}}{2017}]%
        {gollapalli2017incorporating}
\bibfield{author}{\bibinfo{person}{Sujatha~Das Gollapalli},
  \bibinfo{person}{Xiao-Li Li}, {and} \bibinfo{person}{Peng Yang}.}
  \bibinfo{year}{2017}\natexlab{}.
\newblock \showarticletitle{Incorporating Expert Knowledge into Keyphrase
  Extraction}. In \bibinfo{booktitle}{\emph{AAAI}}.
\newblock


\bibitem[\protect\citeauthoryear{Hu and Liu}{Hu and Liu}{2004}]%
        {HuL04}
\bibfield{author}{\bibinfo{person}{Minqing Hu} {and} \bibinfo{person}{Bing
  Liu}.} \bibinfo{year}{2004}\natexlab{}.
\newblock \showarticletitle{Mining and summarizing customer reviews}. In
  \bibinfo{booktitle}{\emph{SIGKDD}}. \bibinfo{publisher}{{ACM}},
  \bibinfo{pages}{168--177}.
\newblock


\bibitem[\protect\citeauthoryear{Kingma and Ba}{Kingma and Ba}{2015}]%
        {KingmaB14}
\bibfield{author}{\bibinfo{person}{Diederik~P. Kingma} {and}
  \bibinfo{person}{Jimmy Ba}.} \bibinfo{year}{2015}\natexlab{}.
\newblock \showarticletitle{Adam: {A} Method for Stochastic Optimization}. In
  \bibinfo{booktitle}{\emph{ICLR}}.
\newblock


\bibitem[\protect\citeauthoryear{Le, Nguyen, and Shimazu}{Le
  et~al\mbox{.}}{2016}]%
        {LeNS16}
\bibfield{author}{\bibinfo{person}{Tho Thi~Ngoc Le}, \bibinfo{person}{Minh~Le
  Nguyen}, {and} \bibinfo{person}{Akira Shimazu}.}
  \bibinfo{year}{2016}\natexlab{}.
\newblock \showarticletitle{Unsupervised Keyphrase Extraction: Introducing New
  Kinds of Words to Keyphrases}. In \bibinfo{booktitle}{\emph{{AI} 2016}}
  \emph{(\bibinfo{series}{LNCS})}, Vol.~\bibinfo{volume}{9992}.
  \bibinfo{pages}{665--671}.
\newblock


\bibitem[\protect\citeauthoryear{Li, Galley, Brockett, Gao, and Dolan}{Li
  et~al\mbox{.}}{2016}]%
        {LiGBGD16}
\bibfield{author}{\bibinfo{person}{Jiwei Li}, \bibinfo{person}{Michel Galley},
  \bibinfo{person}{Chris Brockett}, \bibinfo{person}{Jianfeng Gao}, {and}
  \bibinfo{person}{Bill Dolan}.} \bibinfo{year}{2016}\natexlab{}.
\newblock \showarticletitle{A Diversity-Promoting Objective Function for Neural
  Conversation Models}. In \bibinfo{booktitle}{\emph{NAACL}}.
  \bibinfo{pages}{110--119}.
\newblock


\bibitem[\protect\citeauthoryear{Li, Wang, Yin, and Zong}{Li
  et~al\mbox{.}}{2019b}]%
        {LiWYZ19}
\bibfield{author}{\bibinfo{person}{Junjie Li}, \bibinfo{person}{Xuepeng Wang},
  \bibinfo{person}{Dawei Yin}, {and} \bibinfo{person}{Chengqing Zong}.}
  \bibinfo{year}{2019}\natexlab{b}.
\newblock \showarticletitle{Attribute-aware Sequence Network for Review
  Summarization}. In \bibinfo{booktitle}{\emph{EMNLP-IJCNLP}}.
  \bibinfo{pages}{2998--3008}.
\newblock


\bibitem[\protect\citeauthoryear{Li, Huang, and Ren}{Li et~al\mbox{.}}{2020}]%
        {Lihr19}
\bibfield{author}{\bibinfo{person}{Pengyuan Li}, \bibinfo{person}{Lei Huang},
  {and} \bibinfo{person}{Guang{-}jie Ren}.} \bibinfo{year}{2020}\natexlab{}.
\newblock \showarticletitle{Topic Detection and Summarization of User Reviews}.
\newblock \bibinfo{journal}{\emph{CoRR}}  \bibinfo{volume}{abs/2006.00148}
  (\bibinfo{year}{2020}).
\newblock


\bibitem[\protect\citeauthoryear{Li, Wang, Bing, and Lam}{Li
  et~al\mbox{.}}{2019a}]%
        {li2019persona}
\bibfield{author}{\bibinfo{person}{Piji Li}, \bibinfo{person}{Zihao Wang},
  \bibinfo{person}{Lidong Bing}, {and} \bibinfo{person}{Wai Lam}.}
  \bibinfo{year}{2019}\natexlab{a}.
\newblock \showarticletitle{Persona-Aware Tips Generation}. In
  \bibinfo{booktitle}{\emph{The World Wide Web Conference}}.
  \bibinfo{pages}{1006--1016}.
\newblock


\bibitem[\protect\citeauthoryear{Li, Wang, Ren, Bing, and Lam}{Li
  et~al\mbox{.}}{2017}]%
        {li2017neural}
\bibfield{author}{\bibinfo{person}{Piji Li}, \bibinfo{person}{Zihao Wang},
  \bibinfo{person}{Zhaochun Ren}, \bibinfo{person}{Lidong Bing}, {and}
  \bibinfo{person}{Wai Lam}.} \bibinfo{year}{2017}\natexlab{}.
\newblock \showarticletitle{Neural Rating Regression with Abstractive Tips
  Generation for Recommendation}. In \bibinfo{booktitle}{\emph{SIGIR}}.
  \bibinfo{pages}{345--354}.
\newblock


\bibitem[\protect\citeauthoryear{Liu, Lowe, Serban, Noseworthy, Charlin, and
  Pineau}{Liu et~al\mbox{.}}{2016}]%
        {LiuLSNCP16}
\bibfield{author}{\bibinfo{person}{Chia{-}Wei Liu}, \bibinfo{person}{Ryan
  Lowe}, \bibinfo{person}{Iulian Serban}, \bibinfo{person}{Michael Noseworthy},
  \bibinfo{person}{Laurent Charlin}, {and} \bibinfo{person}{Joelle Pineau}.}
  \bibinfo{year}{2016}\natexlab{}.
\newblock \showarticletitle{How {NOT} To Evaluate Your Dialogue System: An
  Empirical Study of Unsupervised Evaluation Metrics for Dialogue Response
  Generation}. In \bibinfo{booktitle}{\emph{EMNLP}}.
  \bibinfo{pages}{2122--2132}.
\newblock


\bibitem[\protect\citeauthoryear{Liu, Gong, Fu, Liu, Yan, Shao, Jiang, Lv, and
  Duan}{Liu et~al\mbox{.}}{2020a}]%
        {liu2020diverse}
\bibfield{author}{\bibinfo{person}{Dayiheng Liu}, \bibinfo{person}{Yeyun Gong},
  \bibinfo{person}{Jie Fu}, \bibinfo{person}{Wei Liu}, \bibinfo{person}{Yu
  Yan}, \bibinfo{person}{Bo Shao}, \bibinfo{person}{Daxin Jiang},
  \bibinfo{person}{Jiancheng Lv}, {and} \bibinfo{person}{Nan Duan}.}
  \bibinfo{year}{2020}\natexlab{a}.
\newblock \showarticletitle{Diverse, Controllable, and Keyphrase-Aware: A
  Corpus and Method for News Multi-Headline Generation}.
\newblock \bibinfo{journal}{\emph{arXiv preprint arXiv:2004.03875}}
  (\bibinfo{year}{2020}).
\newblock


\bibitem[\protect\citeauthoryear{Liu, Lin, and Wang}{Liu
  et~al\mbox{.}}{2020b}]%
        {liu2020keyphrase}
\bibfield{author}{\bibinfo{person}{Rui Liu}, \bibinfo{person}{Zheng Lin}, {and}
  \bibinfo{person}{Weiping Wang}.} \bibinfo{year}{2020}\natexlab{b}.
\newblock \showarticletitle{Keyphrase Prediction With Pre-trained Language
  Model}.
\newblock \bibinfo{journal}{\emph{arXiv preprint arXiv:2004.10462}}
  (\bibinfo{year}{2020}).
\newblock


\bibitem[\protect\citeauthoryear{Lu, Zhai, and Sundaresan}{Lu
  et~al\mbox{.}}{2009}]%
        {LuZS09}
\bibfield{author}{\bibinfo{person}{Yue Lu}, \bibinfo{person}{ChengXiang Zhai},
  {and} \bibinfo{person}{Neel Sundaresan}.} \bibinfo{year}{2009}\natexlab{}.
\newblock \showarticletitle{Rated Aspect Summarization of Short Comments}. In
  \bibinfo{booktitle}{\emph{WWW}}. \bibinfo{pages}{131--140}.
\newblock


\bibitem[\protect\citeauthoryear{Luan, Ostendorf, and Hajishirzi}{Luan
  et~al\mbox{.}}{2017}]%
        {luan-etal-2017-scientific}
\bibfield{author}{\bibinfo{person}{Yi Luan}, \bibinfo{person}{Mari Ostendorf},
  {and} \bibinfo{person}{Hannaneh Hajishirzi}.}
  \bibinfo{year}{2017}\natexlab{}.
\newblock \showarticletitle{Scientific Information Extraction with
  Semi-supervised Neural Tagging}. In \bibinfo{booktitle}{\emph{EMNLP}}.
\newblock


\bibitem[\protect\citeauthoryear{MacQueen}{MacQueen}{1967}]%
        {macqueen1967some}
\bibfield{author}{\bibinfo{person}{James MacQueen}.}
  \bibinfo{year}{1967}\natexlab{}.
\newblock \showarticletitle{Some Methods for Classification and Analysis of
  Multivariate Observations}. In \bibinfo{booktitle}{\emph{Proceedings of the
  fifth Berkeley symposium on mathematical statistics and probability}},
  Vol.~\bibinfo{volume}{1}. Oakland, CA, USA, \bibinfo{pages}{281--297}.
\newblock


\bibitem[\protect\citeauthoryear{Medelyan, Frank, and Witten}{Medelyan
  et~al\mbox{.}}{2009}]%
        {medelyan2009human}
\bibfield{author}{\bibinfo{person}{Olena Medelyan}, \bibinfo{person}{Eibe
  Frank}, {and} \bibinfo{person}{Ian~H Witten}.}
  \bibinfo{year}{2009}\natexlab{}.
\newblock \showarticletitle{Human-competitive Tagging using Automatic Keyphrase
  Extraction}. In \bibinfo{booktitle}{\emph{EMNLP}}.
  \bibinfo{pages}{1318--1327}.
\newblock


\bibitem[\protect\citeauthoryear{Meng, Zhao, Han, He, Brusilovsky, and
  Chi}{Meng et~al\mbox{.}}{2017}]%
        {MengZHHBC17}
\bibfield{author}{\bibinfo{person}{Rui Meng}, \bibinfo{person}{Sanqiang Zhao},
  \bibinfo{person}{Shuguang Han}, \bibinfo{person}{Daqing He},
  \bibinfo{person}{Peter Brusilovsky}, {and} \bibinfo{person}{Yu Chi}.}
  \bibinfo{year}{2017}\natexlab{}.
\newblock \showarticletitle{Deep Keyphrase Generation}.
\newblock \bibinfo{journal}{\emph{CoRR}}  \bibinfo{volume}{abs/1704.06879}
  (\bibinfo{year}{2017}).
\newblock


\bibitem[\protect\citeauthoryear{Mihalcea and Tarau}{Mihalcea and
  Tarau}{2004}]%
        {mihalcea2004textrank}
\bibfield{author}{\bibinfo{person}{Rada Mihalcea} {and} \bibinfo{person}{Paul
  Tarau}.} \bibinfo{year}{2004}\natexlab{}.
\newblock \showarticletitle{Textrank: Bringing order into text}. In
  \bibinfo{booktitle}{\emph{EMNLP}}. \bibinfo{pages}{404--411}.
\newblock


\bibitem[\protect\citeauthoryear{Novgorodov, Guy, Elad, and
  Radinsky}{Novgorodov et~al\mbox{.}}{2019}]%
        {novgorodov2019generating}
\bibfield{author}{\bibinfo{person}{Slava Novgorodov}, \bibinfo{person}{Ido
  Guy}, \bibinfo{person}{Guy Elad}, {and} \bibinfo{person}{Kira Radinsky}.}
  \bibinfo{year}{2019}\natexlab{}.
\newblock \showarticletitle{Generating Product Descriptions from User Reviews}.
  In \bibinfo{booktitle}{\emph{WWW}}. \bibinfo{pages}{1354--1364}.
\newblock


\bibitem[\protect\citeauthoryear{Ray~Chowdhury, Caragea, and
  Caragea}{Ray~Chowdhury et~al\mbox{.}}{2019}]%
        {ray2019keyphrase}
\bibfield{author}{\bibinfo{person}{Jishnu Ray~Chowdhury},
  \bibinfo{person}{Cornelia Caragea}, {and} \bibinfo{person}{Doina Caragea}.}
  \bibinfo{year}{2019}\natexlab{}.
\newblock \showarticletitle{Keyphrase extraction from disaster-related tweets}.
  In \bibinfo{booktitle}{\emph{WWW}}. \bibinfo{pages}{1555--1566}.
\newblock


\bibitem[\protect\citeauthoryear{See, Liu, and Manning}{See
  et~al\mbox{.}}{2017}]%
        {SeeLM17}
\bibfield{author}{\bibinfo{person}{Abigail See}, \bibinfo{person}{Peter~J.
  Liu}, {and} \bibinfo{person}{Christopher~D. Manning}.}
  \bibinfo{year}{2017}\natexlab{}.
\newblock \showarticletitle{Get To The Point: Summarization with
  Pointer-Generator Networks}. In \bibinfo{booktitle}{\emph{ACL}}.
  \bibinfo{pages}{1073--1083}.
\newblock


\bibitem[\protect\citeauthoryear{Shen, Wang, Wang, Min, Su, Zhang, Li, Henao,
  and Carin}{Shen et~al\mbox{.}}{2018}]%
        {HenaoLCSSWWMZ18}
\bibfield{author}{\bibinfo{person}{Dinghan Shen}, \bibinfo{person}{Guoyin
  Wang}, \bibinfo{person}{Wenlin Wang}, \bibinfo{person}{Martin~Renqiang Min},
  \bibinfo{person}{Qinliang Su}, \bibinfo{person}{Yizhe Zhang},
  \bibinfo{person}{Chunyuan Li}, \bibinfo{person}{Ricardo Henao}, {and}
  \bibinfo{person}{Lawrence Carin}.} \bibinfo{year}{2018}\natexlab{}.
\newblock \showarticletitle{Baseline Needs More Love: On Simple
  Word-Embedding-Based Models and Associated Pooling Mechanisms}. In
  \bibinfo{booktitle}{\emph{ACL}}. \bibinfo{pages}{440--450}.
\newblock


\bibitem[\protect\citeauthoryear{Srivastava, Hinton, Krizhevsky, Sutskever, and
  Salakhutdinov}{Srivastava et~al\mbox{.}}{2014}]%
        {SrivastavaHKSS14}
\bibfield{author}{\bibinfo{person}{Nitish Srivastava},
  \bibinfo{person}{Geoffrey~E. Hinton}, \bibinfo{person}{Alex Krizhevsky},
  \bibinfo{person}{Ilya Sutskever}, {and} \bibinfo{person}{Ruslan
  Salakhutdinov}.} \bibinfo{year}{2014}\natexlab{}.
\newblock \showarticletitle{Dropout: A Simple Way to Prevent Neural Networks
  from Overfitting}.
\newblock \bibinfo{journal}{\emph{J. Mach. Learn. Res.}} \bibinfo{volume}{15},
  \bibinfo{number}{1} (\bibinfo{year}{2014}), \bibinfo{pages}{1929--1958}.
\newblock


\bibitem[\protect\citeauthoryear{Suhara, Wang, Angelidis, and Tan}{Suhara
  et~al\mbox{.}}{2020}]%
        {SuharaWAT20}
\bibfield{author}{\bibinfo{person}{Yoshihiko Suhara}, \bibinfo{person}{Xiaolan
  Wang}, \bibinfo{person}{Stefanos Angelidis}, {and}
  \bibinfo{person}{Wang{-}Chiew Tan}.} \bibinfo{year}{2020}\natexlab{}.
\newblock \showarticletitle{OpinionDigest: {A} Simple Framework for Opinion
  Summarization}. In \bibinfo{booktitle}{\emph{ACL}}.
  \bibinfo{pages}{5789--5798}.
\newblock


\bibitem[\protect\citeauthoryear{Sun, Tang, Du, Deng, and Nie}{Sun
  et~al\mbox{.}}{2019}]%
        {SunTDDN19}
\bibfield{author}{\bibinfo{person}{Zhiqing Sun}, \bibinfo{person}{Jian Tang},
  \bibinfo{person}{Pan Du}, \bibinfo{person}{Zhi{-}Hong Deng}, {and}
  \bibinfo{person}{Jian{-}Yun Nie}.} \bibinfo{year}{2019}\natexlab{}.
\newblock \showarticletitle{DivGraphPointer: {A} Graph Pointer Network for
  Extracting Diverse Keyphrases}. In \bibinfo{booktitle}{\emph{SIGIR}}.
  \bibinfo{pages}{755--764}.
\newblock


\bibitem[\protect\citeauthoryear{Szegedy, Vanhoucke, Ioffe, Shlens, and
  Wojna}{Szegedy et~al\mbox{.}}{2016}]%
        {SzegedyVISW16}
\bibfield{author}{\bibinfo{person}{Christian Szegedy}, \bibinfo{person}{Vincent
  Vanhoucke}, \bibinfo{person}{Sergey Ioffe}, \bibinfo{person}{Jonathon
  Shlens}, {and} \bibinfo{person}{Zbigniew Wojna}.}
  \bibinfo{year}{2016}\natexlab{}.
\newblock \showarticletitle{Rethinking the Inception Architecture for Computer
  Vision}. In \bibinfo{booktitle}{\emph{CVPR}}. \bibinfo{pages}{2818--2826}.
\newblock


\bibitem[\protect\citeauthoryear{Tang, Qin, Liu, and Yang}{Tang
  et~al\mbox{.}}{2015}]%
        {tang2015user}
\bibfield{author}{\bibinfo{person}{Duyu Tang}, \bibinfo{person}{Bing Qin},
  \bibinfo{person}{Ting Liu}, {and} \bibinfo{person}{Yuekui Yang}.}
  \bibinfo{year}{2015}\natexlab{}.
\newblock \showarticletitle{User Modeling with Neural Network for Review Rating
  Prediction}. In \bibinfo{booktitle}{\emph{IJCAI}}.
\newblock


\bibitem[\protect\citeauthoryear{Tay}{Tay}{2019}]%
        {Tay19}
\bibfield{author}{\bibinfo{person}{Wenyi Tay}.}
  \bibinfo{year}{2019}\natexlab{}.
\newblock \showarticletitle{Not All Reviews Are Equal: Towards Addressing
  Reviewer Biases for Opinion Summarization}. In
  \bibinfo{booktitle}{\emph{ACL}}. \bibinfo{pages}{34--42}.
\newblock


\bibitem[\protect\citeauthoryear{Vaswani, Shazeer, Parmar, Uszkoreit, Jones,
  Gomez, Kaiser, and Polosukhin}{Vaswani et~al\mbox{.}}{2017}]%
        {vaswani2017attention}
\bibfield{author}{\bibinfo{person}{Ashish Vaswani}, \bibinfo{person}{Noam
  Shazeer}, \bibinfo{person}{Niki Parmar}, \bibinfo{person}{Jakob Uszkoreit},
  \bibinfo{person}{Llion Jones}, \bibinfo{person}{Aidan~N Gomez},
  \bibinfo{person}{{\L}ukasz Kaiser}, {and} \bibinfo{person}{Illia
  Polosukhin}.} \bibinfo{year}{2017}\natexlab{}.
\newblock \showarticletitle{Attention is all you need}. In
  \bibinfo{booktitle}{\emph{NeurIPS}}. \bibinfo{pages}{6000--6010}.
\newblock


\bibitem[\protect\citeauthoryear{Wang and Ling}{Wang and Ling}{2016}]%
        {WangL16}
\bibfield{author}{\bibinfo{person}{Lu Wang} {and} \bibinfo{person}{Wang Ling}.}
  \bibinfo{year}{2016}\natexlab{}.
\newblock \showarticletitle{Neural Network-Based Abstract Generation for
  Opinions and Arguments}. In \bibinfo{booktitle}{\emph{NAACL}}.
  \bibinfo{pages}{47--57}.
\newblock


\bibitem[\protect\citeauthoryear{Wang, Zhao, and Huang}{Wang
  et~al\mbox{.}}{2016}]%
        {wang2016ptr}
\bibfield{author}{\bibinfo{person}{Minmei Wang}, \bibinfo{person}{Bo Zhao},
  {and} \bibinfo{person}{Yihua Huang}.} \bibinfo{year}{2016}\natexlab{}.
\newblock \showarticletitle{{PTR}: Phrase-based Topical Ranking for Automatic
  Keyphrase Extraction in Scientific Publications}. In
  \bibinfo{booktitle}{\emph{ICONIP}}. \bibinfo{pages}{120--128}.
\newblock


\bibitem[\protect\citeauthoryear{Wang, Li, Chan, King, Lyu, and Shi}{Wang
  et~al\mbox{.}}{2019a}]%
        {WangLCKLS19}
\bibfield{author}{\bibinfo{person}{Yue Wang}, \bibinfo{person}{Jing Li},
  \bibinfo{person}{Hou~Pong Chan}, \bibinfo{person}{Irwin King},
  \bibinfo{person}{Michael~R. Lyu}, {and} \bibinfo{person}{Shuming Shi}.}
  \bibinfo{year}{2019}\natexlab{a}.
\newblock \showarticletitle{Topic-Aware Neural Keyphrase Generation for Social
  Media Language}. In \bibinfo{booktitle}{\emph{ACL}}.
  \bibinfo{pages}{2516--2526}.
\newblock


\bibitem[\protect\citeauthoryear{Wang, Li, King, Lyu, and Shi}{Wang
  et~al\mbox{.}}{2019b}]%
        {WangLKLS19}
\bibfield{author}{\bibinfo{person}{Yue Wang}, \bibinfo{person}{Jing Li},
  \bibinfo{person}{Irwin King}, \bibinfo{person}{Michael~R. Lyu}, {and}
  \bibinfo{person}{Shuming Shi}.} \bibinfo{year}{2019}\natexlab{b}.
\newblock \showarticletitle{Microblog Hashtag Generation via Encoding
  Conversation Contexts}. In \bibinfo{booktitle}{\emph{NAACL-HLT}}.
  \bibinfo{pages}{1624--1633}.
\newblock


\bibitem[\protect\citeauthoryear{Xiong and Litman}{Xiong and Litman}{2014}]%
        {XiongL14}
\bibfield{author}{\bibinfo{person}{Wenting Xiong} {and}
  \bibinfo{person}{Diane~J. Litman}.} \bibinfo{year}{2014}\natexlab{}.
\newblock \showarticletitle{Empirical Analysis of Exploiting Review Helpfulness
  for Extractive Summarization of Online Reviews}. In
  \bibinfo{booktitle}{\emph{COLING}}. \bibinfo{pages}{1985--1995}.
\newblock


\bibitem[\protect\citeauthoryear{Zhang, Wang, Gong, and Huang}{Zhang
  et~al\mbox{.}}{2016}]%
        {ZhangWGH16}
\bibfield{author}{\bibinfo{person}{Qi Zhang}, \bibinfo{person}{Yang Wang},
  \bibinfo{person}{Yeyun Gong}, {and} \bibinfo{person}{Xuanjing Huang}.}
  \bibinfo{year}{2016}\natexlab{}.
\newblock \showarticletitle{Keyphrase Extraction Using Deep Recurrent Neural
  Networks on Twitter}. In \bibinfo{booktitle}{\emph{EMNLP}}.
  \bibinfo{pages}{836--845}.
\newblock


\bibitem[\protect\citeauthoryear{Zhang, Li, Song, and Zhang}{Zhang
  et~al\mbox{.}}{2018}]%
        {ZhangLSZ18}
\bibfield{author}{\bibinfo{person}{Yingyi Zhang}, \bibinfo{person}{Jing Li},
  \bibinfo{person}{Yan Song}, {and} \bibinfo{person}{Chengzhi Zhang}.}
  \bibinfo{year}{2018}\natexlab{}.
\newblock \showarticletitle{Encoding Conversation Context for Neural Keyphrase
  Extraction from Microblog Posts}. In \bibinfo{booktitle}{\emph{NAACL-HLT}}.
  \bibinfo{pages}{1676--1686}.
\newblock


\end{thebibliography}


\end{document}